%% file: main.tex
\documentclass[review]{elsarticle}

\usepackage{lineno,hyperref,amsmath}
\usepackage{natbib}  
\usepackage{caption} 
\usepackage{amsmath}
\usepackage{amssymb}
\usepackage{algorithmic}
\usepackage{algorithm}
\usepackage{booktabs}
\usepackage{multirow}
\usepackage{makecell}
\usepackage{pifont}
\usepackage[export]{adjustbox}
\usepackage{xcolor}
\usepackage{subfigure}
\modulolinenumbers[5]

\journal{Journal of \LaTeX\ Templates}

\begin{document}

\begin{frontmatter}
\title{Class-Imbalanced Semi-Supervised Learning for \\ Large-Scale Point Cloud Semantic Segmentation via Decoupling Optimization}
\tnotetext[mytitlenote]{Fully documented templates are available in the elsarticle package on \href{http://www.ctan.org/tex-archive/macros/latex/contrib/elsarticle}{CTAN}.}

%
%
%

\author[1,6]{Mengtian Li}
\ead{mtli@stu.ecnu.ecu.cn}
\author[2,5]{Shaohui Lin\corref{cor1}}
\ead{shlin@cs.ecnu.edu.cn}
\author[2]{Zihan Wang}
\author[3]{Yunhang Shen}
\author[4]{Baochang Zhang} 
\author[2]{Lizhuang Ma}
\cortext[cor1]{Corresponding author}
\address[1]{Shanghai University}
\address[6]{Shanghai Engineering Research Center of Motion Picture Special Effects}
\address[2]{East China Normal University}
\address[5]{Key Laboratory of Advanced Theory and Application in Statistics and Data Science, Ministry of Education}
\address[4]{Beihang University}
\address[3]{Tencent Youtu Lab}

\input{latex/abs}

\begin{keyword}
\texttt{}3D Point Cloud\sep Class-imbalanced Learning\sep Semi-Supervised Learning\sep Semantic Segmentation
\MSC[2010] 00-01\sep  99-00
\end{keyword}

\end{frontmatter}

\input{latex/intro}
\input{latex/related_works}
\input{latex/methods}
\input{latex/experiment}
\input{latex/conclusion}

\linenumbers

\bibliography{main.bib}
\end{document}

%% file: latex/abs.tex
\begin{abstract}

%
Semi-supervised learning (SSL), thanks to the significant reduction of data annotation costs, has been an active research topic for large-scale 3D scene understanding.
However, the existing SSL-based methods suffer from severe training bias, mainly due to class imbalance and long-tail distributions of the point cloud data.
As a result, they lead to a biased prediction for the tail class segmentation.
In this paper, we introduce a new decoupling optimization framework, which disentangles  feature representation learning and classifier in an alternative optimization manner to shift the bias decision boundary effectively.
In particular, we first employ two-round pseudo-label generation to select unlabeled points across head-to-tail classes.  We further introduce  multi-class imbalanced focus loss to adaptively pay more attention to feature learning across head-to-tail classes.
We fix the backbone parameters after feature learning and retrain the classifier using ground-truth points to update its parameters.
Extensive experiments demonstrate the effectiveness of our method outperforming previous state-of-the-art methods on both indoor and outdoor 3D point cloud datasets~(\emph{i.e.}, S3DIS, ScanNet-V2, Semantic3D, and SemanticKITTI) using 1\% and 1pt evaluation.

\end{abstract}

%% file: latex/intro.tex
\section{Introduction}
Learning the precise semantic meanings of large-scale 3D point clouds plays a vital role in real-time AI systems\cite{lv2023kss}, such as autonomous driving\cite{li2022paying} and 3D reconstruction\cite{lv2022intrinsic}.
%
Recently, 3D point cloud semantic segmentation has paid more attention to designing architectures and modules, such as point-wise architecture~\cite{qi2017pointnet}\cite{hu2019randla}, voxel-based framework~\cite{3DSparseConvNet} and point-voxel CNN~\cite{liu2019point}\cite{shuai2021backward}.
However, these methods heavily rely on the availability and quantity of point-wise annotations for fully-supervised learning, which are typically labor-intensive and costly. 

\begin{figure*}[t]
\centering
\includegraphics[width=0.99\textwidth]{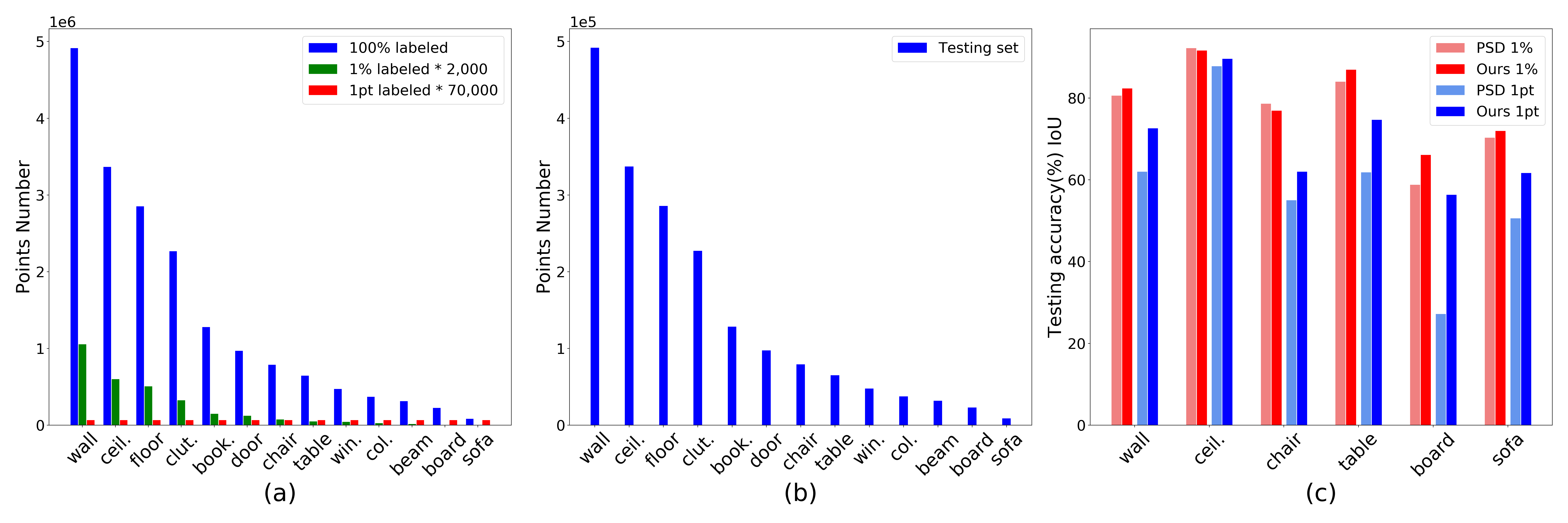}
\caption{Illustration of the widely used S3DIS dataset on training and test setting for class-imbalanced semi-supervised point cloud semantic segmentation.
%
(a) The distribution of annotation data in the \textit{training} set: long-tail distribution of $1\%$ and uniform distribution of 1pt. For a better view of their distributions, the number of labeled points for each is multiplied with the same number, \emph{e.g.}, $2,000$ in $1\%$ and $70,000$ in 1pt.
(b) Long-tail distribution in the \textit{test} set.
(c) IoU of PSD~\cite{zhang2021perturbed} and ours on head \{wall, cell\},  waist \{chair, table\} and tail \{board, sofa\} classes.}
%
%
\label{motivation}
\end{figure*}

To alleviate the annotation burden, previous works have proposed semi-supervised learning~(SSL) for point cloud semantic segmentation to attain the performance of fully-supervised counterparts with a tiny fraction of labeled samples.
For example, PSD~\cite{zhang2021perturbed} provides additional supervision by perturbed self-distillation for implicit information propagation. 
%
1T1C~\cite{liu2021one} proposes a self-training strategy to utilize the pseudo labels to improve the network performance. 
HybridCR~\cite{li2022hybridcr} proposes a novel hybrid contrastive regularization with pseudo labeling.
SQN~\cite{hu2022sqn} leverages a point neighbourhood query to fully utilize the sparse training signals.
LaserMix~\cite{kong2023lasermix} attempt to mix laser beams from different LiDAR scans and then encourage the model to make consistent and confident predictions.
GaIA~\cite{lee2023gaia} aims to reduce the epistemic uncertainty measured by the entropy for a precise semantic segmentation.
%
However, the existing 3D SSL-based methods neglect class-imbalanced problem (\emph{i.e.}, skewed distributions with a long tail) in the real scenarios, which leads to poor SSL performance, especially on long-tail point cloud semantic segmentation~(see PSD~\cite{zhang2021perturbed} on tail classes of ``board'' and ``sofa'' in Fig.~\ref{motivation}).
This is due to the fact that class-imbalanced data can bias the models towards head classes with numerous samples, and away from tail classes with few samples.
%

%
Actually, recent works have proposed re-sampling, re-weighting and transfer learning technologies to balance semi-supervised models for image classification.
For example, CReST~\cite{wei2021crest} re-samples pseudo-labeled samples from tail classes according to the estimated distribution of class frequency.
%
ABC~\cite{lee2021abc} introduces an auxiliary balanced classifier to balance across classes by consistency regularization.
However, these methods are difficult to be adapted to large-scale 3D point cloud semantic segmentation. This is due to the extremely different training and evaluation settings between large-scale point-cloud benchmark datasets and image benchmarks (\emph{e.g.} CIFAR-10~\cite{krizhevsky2009learning}).
On the one hand, labeled and unlabeled training data on images share the same long-tail distribution, while point-cloud datasets (\emph{e.g.}, S3DIS) have different kinds of task settings on labeled points. For example, as shown in Fig.~\ref{motivation}(a), $1$pt and $1\%$ represent only one point and $1\%$ points are randomly labeled for each class, respectively. Therefore, 
the semi-supervised training setting for real point cloud scenarios is more complex, compared to that on images.
%
%
On the other hand, the assumption on testing data distribution is also totally different, i.e., uniform distribution in image benchmark datasets vs. long-tail distribution in point cloud datasets. Under this setting circumstance, off-the-shelf class-imbalance semi-supervised learning~(CISSL) methods~\cite{wei2021crest,lee2021abc} jointly learn representation and classifier,
which do not effectively shift classifier decision boundaries to handle data imbalance and guarantee feature generalization.

\begin{figure*}[t] 
\centering
\includegraphics[width=1.0\textwidth]{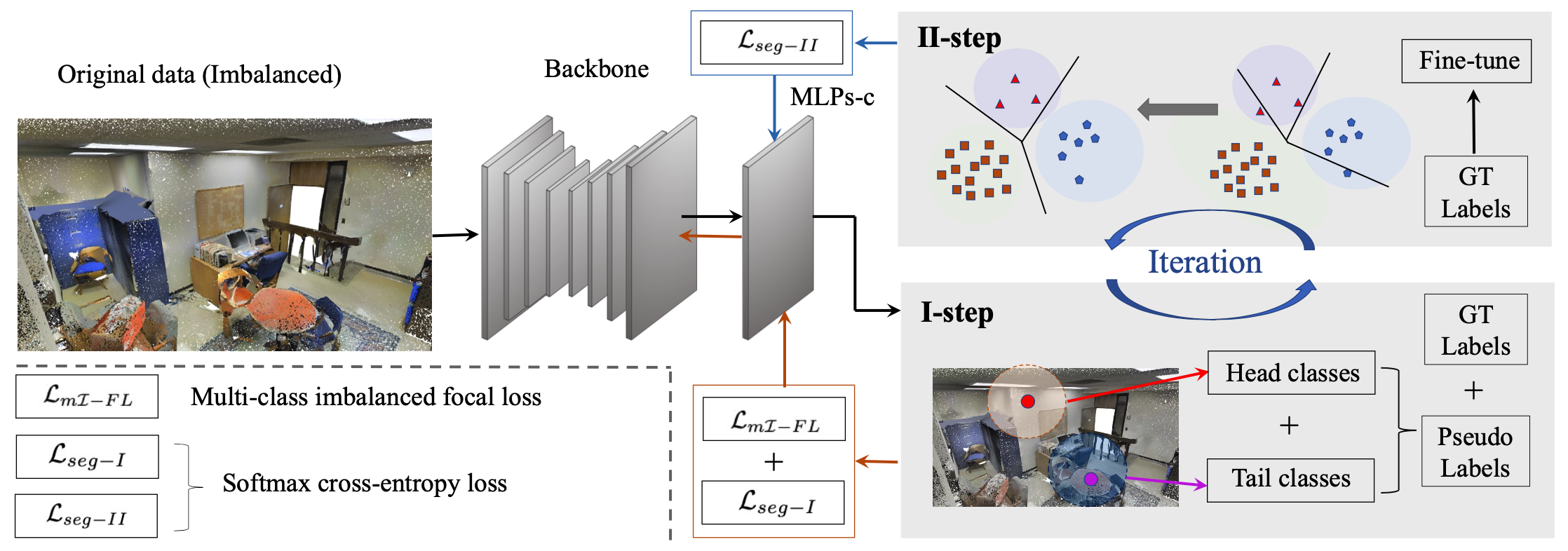} 
\vspace{-15pt}
\caption{The illustration of the decoupling optimization framework. We first pre-train the network with a small number of given labeled points. Then, we conduct alternative optimization to iteratively update the backbone's parameters in the $\uppercase\expandafter{\romannumeral1}-step$ and the classifier's (MLPs-c) parameters in the $\uppercase\expandafter{\romannumeral2}-step$. In particular, two-round pseudo label generation is introduced to sample relative rebalanced points across head-to-tail classes in the $\uppercase\expandafter{\romannumeral1}-step$, which can be used to form multi-class imbalanced focus loss $\mathcal{L}_{m\mathcal{I}-FL}$ for better adaptive feature learning together with ground-truth labeled points by $\mathcal{L}_{seg-I}$. After feature learning, we fine-tune the classifier using the traditional softmax cross-entropy loss $\mathcal{L}_{seg-II}$ on the labeled points.
}
\label{network}
\end{figure*}

%
%
%
%
%
To address the above issues, we propose a new \emph{decoupling optimization} framework for class-imbalanced semi-supervised semantic segmentation on large-scale 3D point clouds. In this framework, we decouple the learning of representation and classifier and alternately update their weights, which better shifts the decision boundaries to separate head-to-tail classes without hurting feature generalization.
%
%
Fig.~\ref{network} depicts the workflow of the proposed approach.
Specifically, we first pre-train the parameters of the backbone and classifier using the available labeled data, and then alternatively update these parameters to learn representation and adjust decision boundary via decoupling optimization.
To better learn the parameters of the backbone, we first employ two-round pseudo label generation to select unlabeled points across head-to-tail classes, where the high-threshold setting in the first round tends to select the high certain points more from head classes and imbalance-rate bootstrap threshold in the second round enables the model to select points from tail classes.
Together with labeled points, we further propose multi-class imbalanced focus loss leveraged into traditional segmentation loss to rebalance head-to-tail point segmentation.
After updating the backbone, we simply retrain the classifier using cross-entropy loss on ground-truth points to update its parameters. 
%


In summary, we make the following contributions: 
\begin{itemize}
\item  To the best of our knowledge, we are the first to propose a decoupling optimization framework for class-imbalanced SSL on large-scale 3D point clouds. It is able to shift bias decision boundaries and learn better feature representation to improve the segmentation performance for class-imbalanced 3D point clouds.
\item  The proposed two-round pseudo-label generation and multi-class imbalanced focus loss are used to adaptively pay attention to feature learning of points from head-to-tail classes.
\item Extensive experimental results demonstrate that our framework achieves new state-of-the-art results and even exceeds its fully supervised counterpart,
e.g., on S3DIS Area-5, we surpass PSD~\cite{zhang2021perturbed} and ~\cite{zhang2021weakly} by $6.8\%$ and $2.4\%$ at $1$pt and $1\%$ settings, respectively.
\end{itemize}

%% file: latex/related_works.tex

\section{Related Work}

\label{related_work}
\noindent
\subsection{Semi-Supervised Point Cloud Segmentation}$\quad$
Existing semi-supervised learning methods can be roughly divided into three categories: 
\textbf{Consistency regularization.} Xu~et al.~\cite{xu2020weakly} introduce a multi-branch supervision method for point cloud feature learning, which adopts two kinds of point cloud augmentation and consistency regularization. Zhang~et al.~\cite{zhang2021perturbed} provide additional supervision by perturbed self-distillation for implicit information propagation, which is implemented by consistency regularization. 
 Shi~et al.~\cite{shi2021label} investigate label-efficient learning and introduce a super-point-based active learning strategy. 
\textbf{Pseudo labeling.} In the semi-supervised setting, Zhang~et al.~\cite{zhang2021weakly} propose a transfer learning-based method and introduce sparse pseudo labels to regularize network learning. 
Hu~et al.~\cite{liu2021one} propose a self-training strategy to utilize the pseudo labels to improve the network performance 
that only requires clicking on one point per instance to indicate its location for annotation which with over-segmentation for pre-processing and extend location annotations into segments as seg-level labels.
 Cheng~et al.~\cite{cheng2021sspc} utilize a dynamic label propagation scheme to generate pseudo labels based on the built super-point graphs. 
 SQN~\cite{hu2022sqn} leverages a point neighbourhood query to fully utilize the sparse training signals.
 LESS~\cite{liu2022less} leverage prototype learning to get more descriptive point embeddings for outdoor LiDAR point clouds scenes.
 LaserMix~\cite{kong2023lasermix} attempt to mix laser beams from different LiDAR scans and then encourage the model to make consistent and confident predictions.
 GaIA~\cite{lee2023gaia} aims to reduce the epistemic uncertainty measured by the entropy for a precise semantic segmentation.
\textbf{Contrastive pre-training.} Xie~et al.~\cite{xie2020pointcontrast} propose a contrastive learning framework for point cloud scenes. However, it mainly focuses on downstream tasks with 100\% labels.
 Hou~et al.~\cite{hou2021exploring} leverage the inherent properties of scenes to expand the network transferability. 
Li~et al.~\cite{jiang2021guided} propose the guided point contrastive loss and leverage pseudo-label to learn discriminative features.
An~et al.~\cite{tao2022seggroup} propose under the assumption of uniform distribution of classes, which cannot well handle the segmentation of points from tail classes in the realistic data-imbalanced case. 
 
\noindent 
\subsection{Class-Imbalanced Supervised Learning}$\quad$
Recent studies on class-imbalanced supervised learning mainly contain three directions: resampling~\cite{zhou2020bbn}\cite{kang2019decoupling},
re-weighting~\cite{cui2019class}\cite{cao2019learning} and transfer learning~\cite{kim2020m2m}\cite{liu2020deep}. 
Re-sampling methods manually sample the data by a pre-defined distribution to get a more balanced training set; Re-weighting methods assign higher weights to tail class instances to balance the overall contribution, while transfer learning aims to transfer knowledge from head classes to tail classes. 
Recent work~\cite{kang2019decoupling} shows that in a decoupled learning scenario, a simple re-sampling strategy can achieve state-of-the-art performance compared to more complicated counterparts. 
However, these methods heavily rely on fully supervised labels, and their performance has not been evaluated extensively under the SSL scenario, especially on large-scale 3D point cloud scenes.

\noindent
\subsection{Class-Imbalanced Semi-Supervised Learning}$\quad$
Recently works have been proposed for imbalanced SSL for image classification. For example, Yang et al.~\cite{yang2020rethinking} find that more accurate decision boundaries can be obtained in class-imbalanced settings through self-supervised learning and semi-supervised learning. DARP~\cite{kim2020distribution} refines biased pseudo labels by solving a convex optimization problem. CReST~\cite{wei2021crest}, a recent self-training technique, mitigates class imbalance by using pseudo-labeled unlabeled data points classified as tail classes with a higher probability than those classified as head classes. ABC~\cite{lee2021abc} introduces an auxiliary balanced classifier of a single layer, which is attached to a representation layer of existing SSL methods.
CoSSL~\cite{fan2022cossl} designs a novel feature enhancement module for the minority class using mixup [41] to train balanced classifiers. Although these algorithms can significantly enhance performance, they assume identical class distributions of labeled and unlabeled data. 
A recent work, DASO~\cite{oh2022daso}, proposes to handle this issue by employing a dynamic combination of linear and semantic pseudo-labels based on the current estimated class distribution of unlabeled data. It is noted that the accuracy of semantic pseudo-labels in DASO relies on the discrimination of learned representations. 
However, these methods assume that the distributions between labeled and unlabeled points are the same, which is totally different from the settings of more complex training in real 3D scenarios. Moreover, different from these methods with joint training, our method proposes decoupling optimization to shift the bias decision boundary better in the more challenging 3D point cloud benchmarks.

%% file: latex/methods.tex
\section{Method}
\label{method}
In this Section, we present the preliminaries and notations in Section \ref{preliminaries}. Then, we introduce our proposed decoupling framework in Section \ref{approaches}, which first initializes the network parameters using the available labeled data. After pre-training, feature representation is learned by the proposed two-round pseudo-label generation and multi-class imbalanced focus loss, while the classifier is updated via simple fine-tuning. 
%
 
\subsection{Preliminaries}
\label{preliminaries}
\textbf{Problem setup and notation.} Let $\mathcal{D}$ be the point cloud dataset, which is defined as $\left\{\left(X^{l}, Y^{l}\right),\left(X^{u}, \varnothing\right)\right\} = \left\{\left(x_{1}^{l}, y_{1}^{l}\right), \ldots,\left(x_{M}^{l}, y_{M}^{l}\right), x_{M+1}^{u}, \ldots, x_{N}^{u}\right\}$, where $N$ and $M$ are the total number of points and the number of labeled points, respectively; $X^{l}$ and $X^{u}$ are the sets of the labeled and unlabeled points, respectively.  
The number of training examples in $X^{l}$ belonging to class $c$ is denoted as $M_{c}$, \emph{i.e.}, $\sum_{c=1}^{C} M_{c}=M$. $\mathcal{D}$ is a long-tail distribution across all classes if $M=N$. 
We assume that the classes are sorted by cardinality in descending order, \emph{i.e.}, $M_{1} \geq M_{2} \geq \cdots \geq M_{C}$ when using the same percentage labeling (\emph{e.g.} 1\%) in $\mathcal{D}$. The marginal class distribution of $X^{l}$ is skewed, \emph{i.e.}, $M_{1} \gg M_{C}$. Since we have two different labeling settings for semi-supervised semantic segmentation, $X^{l}$ and $X^{u}$ do not necessarily share the same distributions. We take $1\%$ and 1pt settings for example.  
For $1\%$ setting, the number of labeled points is $M=1 \% \times N$, and labeled set $X^{l}$ and unlabeled set $X^{u}$ share the same long-tail distribution. 1pt setting represents only one labeled point for each class, where the number of labeled points $M$ equals the number of classes $C$, and labeled set $X^{l}$ is uniformly distributed different to the long-tail distribution in unlabeled set $X^{u}$. 
Note that all labeled points are selected randomly. 
For better discussion, we neglect $1$pt setting satisfying with $M_{1}=M_{2}=\cdots=M_{C}$, as our method also works well in this setting.

For $X^{u}$, the labels are absent and are often replaced by pseudo labels $\hat{Y}$ generated on the fly, and $P$ is the number of the pseudo labels, where $ P+M \leq N$. Thus, $Y= Y^{l} \cup \hat{Y}$ are the whole label sets for weakly-supervised semantic segmentation.
Note that $Y^{l}$ is fixed, but $\hat{Y}$ is updated during training. 
Formally, weakly-supervised semantic segmentation aims to learn the function: $f_{\theta}: X^{l} \cup X^{u} \mapsto Y$, 
where $\theta =  \theta_{b} \cup \theta_{cls}$,  $\theta_{b}$ and $\theta_{cls}$ are the parameters of backbone and classifier, respectively. We denote $\mathbf{Y}^l$ and $\mathbf{\hat{Y}}$ as the final probability outputs of $f_{\theta}(\mathbf{X}^l)$ and $f_{\theta}(\mathbf{X}^u)$, respectively. 
For the testing data, point clouds are class-imbalanced, which is different from the class-balanced test set in the image domain~\cite{wei2021crest,lee2021abc,kim2020distribution}.
%
%
%
%
To fully use the unlabeled points, we formulate the loss as the weighted combination of supervised and unsupervised loss with network parameters $\theta$:
\begin{equation}
\theta^{*}=\underset{\theta}{\arg \min }\sum_{i=1}^{M}-y_i^l\log f_{\theta}(x_i^l) + \lambda\sum_{i=M+1}^{M+P}-\hat{y}_i\log f_{\theta}(x_i^u).
\label{eq1}
\end{equation}

To solve Eq.~\ref{eq1}, previous methods~\cite{zhang2021perturbed} directly employ joint learning for backbone parameters $\theta_b$ and classifier parameters $\theta_{cls}$, which cannot effectively shift bias decision boundary in the class-imbalanced situation. The main reason is that 
(1) Coupled optimization strategy. Joint training with $\theta_b$ and $\theta_{cls}$ significantly reduces the decision area for tail classes, which leads to more biased decision boundaries. 
(2) Insufficient pseudo-label generation. The pseudo labels generated by previous methods~\cite{liu2021one} have a high probability to select points from head classes, such that imbalanced training affects the performance for point cloud segmentation. 
Therefore, we design a decoupling optimization strategy to effectively shift bias decision boundary and construct effective pseudo labels and new focus loss for re-balancing head-to-tail points, such that the model adaptively pays attention to point feature learning from head-to-tail classes. 

\subsection{Decoupling Optimization}
\label{approaches}
\subsubsection{\uppercase\expandafter{\romannumeral1}-step: Fixed classifier $\theta_{cls}$, backbone optimization for 3D feature learning.}
After the warm-up, we generate pseudo labels $\hat{Y}$ for unlabeled data $X^{u}$. The pseudo labels set $\hat{Y}=\left\{\left(x_{i}, \hat{{y}_{i}}\right)\right\}_{i=1}^{N-M}$ is added into the labeled set, \emph{i.e.}, $Y^{\prime}= Y^{l} \cup \hat{Y}, Y^{\prime} \subseteq Y$ for next generation. In large-scale 3D scenarios, it is necessary to generate more pseudo labels on tail classes to ease the imbalance problem. Motivated by this, we propose a dynamic strategy to generate pseudo labels according to the imbalanced ratio of class. 

\textbf{Pseudo label generation.} We first use a moving window threshold to eliminate fluctuations of predictions in different sub-point clouds and reduce false predictions, instead of the fixed threshold.
Let $\hat{\boldsymbol{Y}} \in \mathbb{R}^{N^{\prime} \times C}$ be the final normalized probability results. 
For class $c$, we choose the moving threshold $\delta_{c}^{\mathrm{cer}}$ to select the pseudo label set and donate it as the certain one as $\hat{Y}$ by:
\begin{equation}
\label{eq_pseudo}
\delta_{c}^{\mathrm{cer}}=\max \left(\max _{i}\left(\hat{\boldsymbol{Y}}_{ic}\right)-\delta_{len}, \delta_{d} \right), 
\end{equation}
where $\delta_{len}$ represents the width of the threshold window, $\delta_{d}$ is denoted as a
a lower bound of the threshold, which is set to greater than $0.5$. Then, for each $x_{i}$, we can get the pre-select pseudo label $\hat{y}_{i} = \left[\hat{y}_{i 1}, \ldots, \hat{y}_{i C} \right]$, where $\hat{y}_{i c}=\mathbb{1}\left[\hat{\boldsymbol{Y}}_{i c}>\delta_{c}^{\mathrm{cer}} \right]$.
However, this straightforward way may still be biased towards dominant and overly head classes, which ignores tail classes resulting in more serious imbalance problems.
Thus, we expand the pseudo label set with a selected subset $\hat{S}$ from the rest of the uncertain labels. 
We choose $\hat{S}$ following a class-rebalancing rule inspired by CReST~\cite{wei2021crest}: the less frequent a class $c$ is, the more unlabeled samples that are predicted as class $c$ could hold high precision.
Specifically, at each iteration, we rank the number of predicted labels of each class and then obtain the tail classes for each remaining unlabeled point $x_{j}$, which is predicted as tail classes are added into $\hat{S}$ at the enlarge threshold $\delta_{c}^{\mathrm{uncer}}$:
\begin{equation}
\label{eq_resample}
\delta_{c}^{\mathrm{uncer}}=\min \left( \max _{j}\left(\hat{\boldsymbol{Y}}_{jc}\right),  \left( \frac{1}{ \rho_{c}} \right)^{\beta}  \right),
\end{equation}
where $0 \leq \beta \leq 1 $ tunes the threshold rate and thus the size of $\hat{S}$, $\rho_{c}= \frac{M_{c}}{M_{C}}$  indicates the imbalanced ratio of the $c$-th class.
For $\beta=1$, the $\delta_{c}^{\mathrm{uncer}}$ is more tolerant for the tail class according to its smaller $\rho_{c}$.
For $\beta=0$ (i.e., $\left(\frac{1}{\rho_{c}}\right)^{\beta}=1$) for all class $c$, all uncertain labels are ignored.
By using Eq.~\ref{eq_resample}, we obtain the pre-select pseudo label $\hat{y}_{j } = \left[\hat{y}_{j 1}, \ldots, \hat{y}_{j C} \right]$, where $\hat{y}_{jc} =\mathbb{1}\left[\hat{\boldsymbol{Y}}_{j c}>\delta_{c}^{\mathrm{uncer}} \right]$. To this end, we construct the pseudo label set as $\hat{Y} = \hat{Y} \cup \hat{S}$. 

\textbf{Backbone's parameters updating.} For large-scale 3D point scenes, the head class could provide more geometry features, which dominates the feature learning process to generate a biased model. We attempt to alleviate this imbalanced issue by designing a novel loss to guide and correct the biased model.  
Focal loss~\cite{lin2017focal} is the widely-used solution to the foreground-background imbalance problem in dense object detection, which can be formulated as:
\begin{equation}
\mathcal{L}_\mathrm{FL}=-\alpha\left(1-p_{\mathrm{t}}\right)^{\gamma} \log \left(p_{\mathrm{t}}\right),
\label{fl}
\end{equation}
where $p_{\mathrm{t}} \in[0,1]$ indicates the predicted confidence score of an object candidate, $\alpha$ is the parameter that balances the importance of the samples, and $\gamma$ is the focusing parameter. We expand the focal loss on the segmentation tasks, as Eq.~\ref{fl} is for binary classification. 
%
Therefore, we reformulate the focal loss to multi-class counterpart in imbalanced-SSL segmentation task as:
\begin{equation}
\label{eq_xfl}
\mathcal{L}_{m\mathcal{I}-\mathrm{FL}}=-\sum_{i=1}^{|\{\hat{Y}\}|}\sum_{c=1}^{C} \alpha_{\mathrm{t}}\left(1-p_{\mathrm{t},i}\right)^{ \gamma^{c}} \log \left(p_{\mathrm{t},i}\right),
\end{equation}
where $\alpha_{\mathrm{t}}$ represents the predicted confidence scores for points, which is set to $0.5$; $|\{\hat{Y}\}|$ is the number of points with pseudo labels; $p_{\mathrm{t}, i}=f_{\theta}(x_i^u, \theta|\hat{y}_i\in \hat{Y})$ is the probability of points. $\gamma^{c}$ is the focusing factor for the $c$-th class, which plays a vital role in the imbalance degree of the $c$-th class in the large-scale 3D scenario. 
%
%
Naturally, we adopt a large $\gamma^{c}$ to alleviate the severe imbalance issue in the tail classes, while a small $\gamma^{c}$ is for head classes.
Moreover, inspired by EQLv2~\cite{tan2021equalization}, we also introduce the gradient-guided mechanism to choose $\gamma^{c}$ for balancing the training process of each sample independently and equally. Therefore, focusing factor $\gamma^{c}$ contains two components including a class imbalanced ratio $\rho_{c}$ and a class-specific component $s\left(1-g^{c}\right)$, which is formulated as:
\begin{equation}
\label{gamma}
\gamma^{c} = s\left(1-g^{c}\right) - \frac{1}{ \rho_{c}},
\end{equation}
where $\rho_{c}= \frac{M_{c}}{M_{C}}$ indicates the imbalanced ratio of the $c$-th class, which decides the basic behavior of the classifier.
The hyper-parameter $s$ is a scaling factor that determines the upper limit of $\gamma^{c}$.
%
Parameter $g^{c}$ indicates the accumulated gradient ratio of the $c$-th class. Large $g^{c}$ indicates that the $c$-th class~(\textit{a.k.a.} head classes) is trained in a balanced way, while small one means the class~(\textit{a.k.a.} tail classes) is trained  in an imbalanced way. To satisfy our requirement about the $\gamma^{c}$, we set $g^{c} \in[0,1]$ and let $1-g^{c}$ to invert the distribution.

Compared with $\mathcal{L}_{\mathrm{FL}}$, $\mathcal{L}_{m\mathcal{I}-\mathrm{FL}}$ handles the imbalance problem of each class independently, which leads to a significant performance improvement on tail classes (\emph{cf}. Sec.~\ref{ab_mxfl} for more detailed analysis). 
Therefore, the learning of backbone parameters is constructed by minimizing the following equation:
\begin{equation}
\label{eq_seg}
\mathcal{L}_{\mathrm{fea}} = \mathcal{L}_{m\mathcal{I}-\mathrm{FL}} + \mathcal{L}_{\mathrm{seg}-\uppercase\expandafter{\romannumeral1}}
,
\end{equation} 
where $\mathcal{L}_{m\mathcal{I}-\mathrm{FL}}$ is given in Eq.~\ref{eq_xfl}. The softmax cross-entropy loss is employed as a basic part to promote the performance of the segmentation task. We utilize the softmax cross-entropy loss of labeled points as:
\begin{equation}
\label{step1_seg}
\mathcal{L}_{\mathrm{seg}-\uppercase\expandafter{\romannumeral1}}=-\frac{1}{P+M} \sum_{i=1}^{P+M} \sum_{c=1}^{C} \boldsymbol{y}_{i c} \log \frac{\exp \left(\boldsymbol{y}^{\prime}_{i c}\right)}{\sum_{c=1}^{C} \exp \left(\boldsymbol{y}^{\prime}_{i c}\right)}
,
\end{equation} 
where $\boldsymbol{y}_{ic}$ is the corresponding ground truth of labeled point $i$, $\boldsymbol{y}^{\prime}_{i c}$is the predictions of the labeled point $i$, and $P$ is the number of pseudo labels.
%
\subsubsection{\uppercase\expandafter{\romannumeral2}-step: Fix backbone's parameters, update classifier $\theta_{cls}$.}
This step aims to fine-tune the classifier by fixing the parameters of the backbone optimized. The classifier is regarded as to re-balance the prediction for classes which is imbalanced in large-scale 3D scenarios. To accommodate this, softmax cross-entropy loss conducted on the ground-truth labeled data, data with pseudo labels, or mixed data with labeled and pseudo labels are used to retrain the classifier. We only choose the $Y^{l}$ to fine-tune the classifier in our framework for the reason of its competitive performances and efficiency during computation, without reloading the pseudo labels once again (\emph{cf}. Sec.~\ref{ab_labelsetting} for more detailed analysis). Therefore, the classifier $\theta_{cls}$ is optimized by minimizing the following formulation:
\begin{equation}
\mathcal{L}_{\mathrm{seg}-\uppercase\expandafter{\romannumeral2}}=-\frac{1}{M}\sum_{i=1}^{M} \sum_{c=1}^{C} \boldsymbol{y}_{ic}^{l} \log \frac{\exp \left(\boldsymbol{y}^{l}_{ic}\right)}{\sum_{c=1}^{C} \exp \left(\boldsymbol{y}^{l}_{ic}\right)}.
\label{seg-II}
\end{equation} 

Overall, Alg.~\ref{algorithm_em} presents our decoupling optimization process, which iteratively optimizes \uppercase\expandafter{\romannumeral1}-step and \uppercase\expandafter{\romannumeral2}-step.
\begin{algorithm}[tb]
\algsetup{linenosize=\tiny} 
\caption{Training Procedure for Decoupling Optimization}
\label{algorithm_em}
\begin{algorithmic}
\STATE {\bfseries Input:} $\mathcal{D}=\left\{\left(X^{l}, Y^{l}\right),\left(X^{u}, \varnothing\right)\right\}$; ${\theta} = \{ \theta_{b} , {\theta_{cls} } \}$; Iter=10; 
\romannumeral2-epoch=100; \romannumeral1-epoch=30.
\STATE {\bfseries Output:} $\theta =  \{ \theta_{b} , \theta_{cls} \}$. 
\STATE {\bfseries Pre-train:} Initializing the network parameters ${\theta}$  with $X^{l} \cup X^{u}$ \\
\FOR{$i=0$ {\bfseries to} Iter}
\STATE {\bfseries \uppercase\expandafter{\romannumeral1}-step: repeat \romannumeral1-epoch}\\
Generate pseudo labels with Eq.\ref{eq_pseudo} and Eq.\ref{eq_resample}.\\
Optimize ${\theta_{b}}$ with pseudo labels and ground-truth labels by minimizing Eq.\ref{eq_seg}.
\STATE {\bfseries \uppercase\expandafter{\romannumeral2}-step: repeat \romannumeral2-epoch}\\ 
Fine-tune ${\theta_{cls}}$ with softmax cross-entropy loss via Eq.~\ref{seg-II}.
\ENDFOR
\end{algorithmic}
\end{algorithm}

%% file: latex/experiment.tex
\section{Experiments}
\label{experiments}
\subsection{Experiment setting}
\textbf{Datasets.} We evaluate our method on four widely-used benchmark datasets for point cloud semantic segmentation, S3DIS~\cite{armeni20163d}, ScanNet-V2~\cite{dai2017scannet}, Semantic3D~\cite{hackel2017semantic3d} and SemanticKITTI~\cite{behley2019semantickitti}. 
S3DIS has $271$ point cloud indoor scenes across $6$ areas with $13$ classes, which is split into a training set (Area 1,2,3,4,6) and a validation set (Area 5).
ScanNet-V2 contains $1,613$ 3D indoor scans with $20$ classes, which are split into a training set of $1,201$ scans, a validation set of 312 scans, and a testing set of $100$ scans.
Semantic3D provides over $4$ billion points covering diverse outdoor urban scenes and with $8$ classes, which contain a training set of $15$ scenes, a validation set of $2$ scenes, and a $reduced$-$8$ testing of $4$ scenes.
SemanticKITTI is an outdoor autonomous driving scenario with $19$ classes, which contains $22$ sequences that are divided into a training set of $10$ sequences with $\sim 19$k frames, a validation set of $1$ sequence with $\sim 4$k frames, and a testing set of $11$ sequences with $\sim$20k frames.

\begin{table*}[t]
\centering
\begin{adjustbox}{max width=1.0\linewidth}
\begin{tabular}{c|c|c|ccccccccccccc}
\toprule
Settings & Methods & $m$IoU & ceil. & floor & wall & beam & col. & wind. & door & chair & table & book. & sofa & board & clutter  \\
\midrule
\multirow{6}{*}{Fully}       
& RandLA-Net~\hfill\cite{hu2019randla} & 62.4& 91.2& 95.7& 80.1& 0.0& 25.2& 62.3& 47.4& 75.8& 83.2&  60.8& 70.8& 65.2& 54.0 \\
& RFCR~\hfill\cite{gong2021omni} &  68.7& 94.2& 98.3& 84.3& 0.0& 28.5&  62.4& 71.2&  92.0& 82.6& 76.1& 71.1& 71.6& 61.3 \\
& PSD~\hfill\cite{zhang2021perturbed} & 65.1& 92.3& 97.1& 80.7& 0.0& 32.4& 55.5& 68.1&   78.9& 86.8&  71.1& 70.6& 59.0& 53.0  \\
& HybridCR~\hfill\cite{li2022hybridcr} &  65.8& 93.6& 98.1& 82.3&  0.0& 24.4& 59.5& 66.9& 79.6& 87.9& 67.1& 73.0& 66.8& 55.7 \\
& Ours($ w/o~\mathcal{L}_{m\mathcal{I}-\mathrm{FL}}$)~\hfill{  } & 65.7& 92.3&  97.7&  83.7& 0.0& 22.4& 61.9& 61.8& 77.6&  88.3& 69.2&  72.6& 72.3& 54.1\\
& Ours~\hfill{  }& 66.6&  93.4& 97.4& 83.1& 0.0& 27.2& 63.2& 68.9& 76.5& 88.8& 67.0&  72.6& 72.1& 55.0\\
\midrule
\midrule
\multirow{1}{*}{10\%}                      
& Xu~et al.~\hfill\cite{xu2020weakly} & 48.0& 90.9& 97.3& 74.8& 0.0& 8.4& 49.3& 27.3& 69.0& 71.7& 16.5& 53.2& 23.3& 42.8 \\
\midrule
\multirow{6}{*}{$1\%$}  
& Zhang~et al.~\hfill\cite{zhang2021weakly} & 61.8& 91.5& 96.9& 80.6& 0.0& 18.2& 58.1& 47.2& 75.8& 85.7& 65.3& 68.9& 65.0& 50.2 \\
& PSD~\hfill\cite{zhang2021perturbed} & 63.5& 92.3& 97.7& 80.7& 0.0& 27.8& 56.2& 62.5& 78.7& 84.1& 63.1 & 70.4& 58.9& 53.2 \\
& HybridCR~\hfill\cite{li2022hybridcr} & 65.3& 92.5& 93.9& 82.6& 0.0& 24.2& 64.4& 63.2& 78.3& 81.7& 69.0& 74.4& 68.2& 56.5 \\
& GaIA~\hfill\cite{lee2023gaia}& 66.5& -& -& -& -& -& -& -& -& -& -& -& -& -\\ 
& Ours($ w/o~\mathcal{L}_{m\mathcal{I}-\mathrm{FL}}$)~\hfill{ }& 61.8& 91.0& 95.6& 81.6&  0.0& 22.0& 60.6& 46.2& 75.8& 85.4& 52.0& 70.9& 69.7& 52.3\\
& Ours~\hfill{  }& 68.2& 91.7& 95.5& 82.5& 0.0& 46.6& 63.3& 65.4& 77.0& 89.0& 64.7& 74.5& 69.2& 67.2\\
\midrule
\midrule
\multirow{3}{*}{\makecell{$1$pt \\ (0.2\%) }}                  
& $\Pi$ Model~\hfill\cite{samuli2017temporal} &  44.3& 89.1& 97.0& 71.5& 0.0& 3.6& 43.2& 27.4& 62.1& 63.1& 14.7& 43.7& 24.0& 36.7 \\
& MT~\hfill\cite{tarvainen2017mean} &  
44.4& 88.9& 96.8& 70.1&  0.1& 3.0& 44.3& 28.8& 63.6& 63.7& 15.5& 43.7& 23.0& 35.8 \\
& Xu~et al.~\hfill\cite{xu2020weakly} &  
44.5& 90.1& 97.1& 71.9& 0.0& 1.9& 47.2& 29.3& 62.9& 64.0& 15.9& 42.2& 18.9& 37.5 \\
\midrule
\multirow{6}{*}{\makecell{$1$pt \\ (0.03\%)}} 
& RandlA-Net~\hfill\cite{hu2019randla} & 40.7& 83.7& 90.7& 61.2& 0.0& 11.9& 40.8& 15.2& 52.0& 51.7& 14.9& 50.5& 25.3& 31.8 \\
& PSD~\hfill\cite{zhang2021perturbed} &  48.2& 87.9& 96.0& 62.1& 0.0& 20.6& 49.3& 40.9& 55.1& 61.9& 43.9& 50.7& 27.3& 31.1 \\
& HybridCR~\hfill\cite{li2022hybridcr}  &  51.5& 85.4& 91.9& 65.9&  0.0& 18.0& 51.4& 34.2& 63.8& 78.3& 52.4& 59.6& 29.9& 39.0\\
& GaIA~\hfill\cite{lee2023gaia}& 53.7& -& -& -& -& -& -& -& -& -& -& -& -& -\\ 
&  Ours($ w/o~\mathcal{L}_{m\mathcal{I}-\mathrm{FL}}$)~\hfill{ } & 50.3& 89.8& 96.1& 73.5&  0.0& 22.0& 51.0& 33.7& 54.5& 62.1& 31.4& 59.1& 43.3& 38.0\\
& Ours~\hfill{ }& 55.0& 89.7& 95.5&  72.7& 0.2& 23.5& 52.5& 40.6& 64.1& 78.7& 46.1& 61.8& 50.4& 39.3\\
\bottomrule
\end{tabular}
\end{adjustbox}
\caption{Quantitative results on Area-5 of S3DIS. Note that $1$pt denotes only one labeled point for each class in the entire room instead of small blocks (\emph{e.g.}, $1 \times 1$ meter) of Xu~et al.~\cite{xu2020weakly}. The number of labeled points in our $1$pt setting accounts for $0.03\%$ of the total points, while $0.2\%$ labeled points are used in Xu~et al.~\cite{xu2020weakly}. In the per-class columns, righter classes tend to be more tail in table.
}
\label{s3dis_tab}
\end{table*}
%

\begin{table*}[htp]
\begin{center}
\resizebox{\textwidth}{20mm}{
\begin{tabular}{c|l|c|cccccccccccccccccccc}
\toprule
Set. & Methods & \textbf{\rotatebox{90}{mIoU(\%)}} & \rotatebox{90}{wall}& \rotatebox{90}{floor}& \rotatebox{90}{chair}& \rotatebox{90}{door}& \rotatebox{90}{table}& \rotatebox{90}{cabinet}& \rotatebox{90}{other-furniture}& \rotatebox{90}{window}& \rotatebox{90}{bed}& \rotatebox{90}{sofa}&  \rotatebox{90}{bookshelf}& \rotatebox{90}{desk}& \rotatebox{90}{curtain}& \rotatebox{90}{counter}& \rotatebox{90}{refrigerator}& \rotatebox{90}{picture}& \rotatebox{90}{bath-tub}& \rotatebox{90}{toilet}& \rotatebox{90}{shower-curtain}& \rotatebox{90}{sink} \\
\midrule
\multirow{5}{*}{Fully}  
&PCNN~\hfill\cite{atzmon2018point}& 49.8&  75.1& 94.1& 71.1& 35.2& 50.9& 42.0& 32.4& 50.4& 64.4& 52.9& 56.0& 43.6& 41.4& 22.9& 23.8& 15.5& 55.9& 81.3& 38.7& 49.3\\
&SegGCN~\hfill\cite{lei2020seggcn}& 58.9&  77.1& 93.6& 78.9& 48.4& 56.3& 51.4& 39.6& 49.3& 73.1& 70.0& 53.9&  57.3& 46.7& 44.8& 50.1& 6.1& 83.3& 87.4& 50.7& 59.4\\
&PointConv~\hfill\cite{wu2019pointconv}& 66.6& 81.3& 95.3& 82.2& 50.4& 58.8& 64.4& 42.8& 64.2& 75.9& 75.3& 69.9& 56.4& 77.9& 47.5& 58.6& 20.3& 78.1& 90.2& 75.4& 66.1\\
&KPConv~\hfill\cite{thomas2019kpconv}& 68.4& 81.9& 93.5& 81.4& 59.4& 61.4& 64.7& 45.0& 63.2& 75.8&  78.5& 78.4& 60.5& 77.2& 47.3& 58.7& 18.1& 84.7& 88.2& 80.5& 69.0\\
&RFCR~\hfill\cite{gong2021omni}& 70.2& 82.3& 94.7& 81.8& 61.0& 64.6& 67.2& 47.0& 61.1& 74.5& 77.9&  81.3& 62.3& 81.5& 49.3& 59.4& 24.9& 88.9& 89.2& 84.8& 70.5\\
& HybridCR~\hfill\cite{li2022hybridcr}& 59.9& 87.2& 70.7& 68.3& 56.1& 78.4& 46.3& 61.6& 46.5& 45.6& 93.6& 42.7& 20.7& 46.4& 56.7& 53.1& 69.5& 48.0& 71.3& 76.9& 58.4\\
& Ours($ w/o~\mathcal{L}_{m\mathcal{I}-\mathrm{FL}}$)~\hfill& 62.5& 83.0& 69.4& 75.7& 56.3& 77.2& 44.8& 64.7& 52.0& 50.9& 94.9& 43.1& 19.1& 49.6& 61.4& 64.7& 67.2& 53.5& 87.6& 78.3& 57.1\\
& Ours& 68.4& 71.2& 78.4& 78.2& 65.8& 83.5& 49.9& 82.3& 64.1& 59.7& 95.0& 48.7& 28.1& 57.5& 61.9& 64.7& 76.4& 62.0& 87.1& 84.6& 68.8\\
\midrule
\midrule
\multirow{5}{*}{1\%} 
&PSD~\hfill\cite{zhang2021perturbed}& 54.7& 60.9& 93.3& 77.8& 30.4& 57.2& 46.5& 38.7& 50.6& 67.8& 66.9& 65.9& 49.2& 52.8& 38.8& 43.1& 30.7& 57.1& 71.6& 38.2& 52.6\\
& HybridCR~\hfill\cite{li2022hybridcr}& 56.8& 58.9& 65.8& 66.8& 42.3& 80.2& 36.7& 61.2& 58.1& 45.5& 90.1& 47.5& 33.4& 41.0& 37.5& 51.1& 70.5& 60.8& 71.0& 60.1& 57.9\\
& GaIA~\hfill\cite{lee2023gaia}& 65.2& -& -& -& -& -& -& -& -& -& -& -& -& -& -& -& -& -& -& -& -\\ 
& Ours($ w/o~\mathcal{L}_{m\mathcal{I}-\mathrm{FL}}$)~\hfill& 64.1& 77.6& 70.3& 72.1 & 55.7 & 82.6 & 45.1 & 67.2 & 56.3 & 48.3 & 94.3 & 42.5 & 16.2 & 64.4 & 72.6 & 65.9 & 70.9 & 57.2 & 87.5 & 78.6 & 55.9\\
& Ours& 67.0& 81.6& 77.0& 76.8& 65.2& 80.7& 45.1& 74.7& 65.9& 	54.5& 92.4& 47.3& 14.9& 57.1& 81.1& 63.5& 74.6& 62.3& 89.2& 79.4& 57.0\\
\midrule
\midrule
\multirow{5}{*}{1pt} 
&PSD*~\hfill\cite{zhang2021perturbed}& 47.6& 60.9& 83.2& 71.5& 31.9& 45.6& 32.1& 28.5& 57.5& 36.7& 54.5& 57.8& 37.8&  47.6& 29.9& 49.6& 22.4& 57.7& 75.5& 22.9& 48.4 \\
& HybridCR*~\hfill\cite{li2022hybridcr}& 51.6& 67.6& 59.1& 60.9& 44.2& 77.4& 33.5& 59.7& 42.2& 35.7& 93.2& 34.1& 9.4& 29.8& 52.8& 47.3& 67.6& 49.5& 60.2& 72.1& 34.9\\
& GaIA~\hfill\cite{lee2023gaia}& 52.1& -& -& -& -& -& -& -& -& -& -& -& -& -& -& -& -& -& -& -& - \\ 
& Ours($ w/o~\mathcal{L}_{m\mathcal{I}-\mathrm{FL}}$)~\hfill& 53.8& 49.5& 69.3& 64.7& 47.1& 79.3& 30.0& 47.7& 50.5& 35.8& 90.3& 32.7& 8.1& 47.2& 52.9& 44.8& 71.0& 50.9& 74.6& 73.7& 55.4\\
& Ours& 55.7& 73.5& 66.1& 68.6& 49.1& 74.4 & 39.2& 53.9& 45.1& 37.5& 94.6& 37.6& 20.5& 40.3& 35.6& 55.3& 64.3&	49.7& 82.4& 75.6& 51.5\\
\bottomrule&
\end{tabular}}
\end{center}
\caption{Per-class quantitative results on ScanNet-V2~\cite{dai2017scannet}. In the per-class columns, righter classes tend to be more tail in table. *” denotes the results trained by the official codes.}\label{scannet-tab}
\end{table*}
\begin{table*}[htp]
\begin{center}
\resizebox{\textwidth}{21mm}{
\begin{tabular}{c|l|c|ccccccccccccccccccc}
\toprule
Set. & Methods & \textbf{\rotatebox{90}{mIoU(\%)}} & \rotatebox{90}{vegetation}& \rotatebox{90}{road}& \rotatebox{90}{building}& \rotatebox{90}{sidewalk}& \rotatebox{90}{terrain}& \rotatebox{90}{fence}& \rotatebox{90}{car}& \rotatebox{90}{parking}& \rotatebox{90}{trunk}& \rotatebox{90}{other-ground}& \rotatebox{90}{pole}& \rotatebox{90}{other-vehicle}& \rotatebox{90}{truck}& \rotatebox{90}{traffic-sign}& \rotatebox{90}{person}& \rotatebox{90}{motorcycle}& \rotatebox{90}{bicycle}& \rotatebox{90}{bicyclist}&  \rotatebox{90}{motorcyclist}  \\
\midrule
\multirow{5}{*}{Fully}
&RangeNet53++~\hfill\cite{milioto2019rangenet++}& 52.2& 80.5& 91.8& 87.4& 75.2& 64.6& 58.6& 91.4& 65.0& 55.1& 27.8& 47.9& 23.0& 25.7& 55.9& 38.3& 34.4& 25.7& 38.8& 4.8   \\
&RandLA-Net~\hfill\cite{hu2019randla}& 53.9& 81.4& 90.7& 86.9& 73.7& 66.8& 56.3& 94.2& 60.3& 61.3& 20.4& 49.2& 38.9& 40.1&  47.7& 49.2& 25.8& 26.0& 48.2& 7.2  \\
& HybridCR~\hfill\cite{li2022hybridcr}& 54.0& 90.5& 73.9& 59.1& 21.2& 88.3& 93.9& 42.7& 22.8& 31.6& 36.8& 81.7& 61.7& 66.1& 50.2& 45.5& 49.0& 57.4& 49.5& 4.5\\
& Ours($ w/o~\mathcal{L}_{m\mathcal{I}-\mathrm{FL}}$)~\hfill& 56.5& 78.4&  91.1&  87.4& 74.6& 65.3& 57.4& 88.1& 61.9& 61.6& 26.0& 49.9& 30.0& 36.3& 59.2& 57.5& 35.9& 48.6& 56.8& 7.5\\
& Ours& 57.5& 95.2& 34.3& 26.1& 49.1& 42.4& 47.9& 50.8& 22.8& 90.8& 65.8& 75.5& 31.9& 89.8& 65.3& 83.5& 61.5& 67.7& 44.3& 47.9 \\
\midrule
\midrule
\multirow{4}{*}{1\%} 
&PSD*~\hfill\cite{zhang2021perturbed}& 49.9& 78.5& 90.1& 84.4& 73.0& 64.5& 53.1& 88.2& 69.0& 55.9& 28.9& 40.4&  19.3& 13.0& 50.9& 34.1& 38.1& 29.5& 35.8& 1.4 \\
& HybridCR~\hfill\cite{li2022hybridcr}& 52.3& 89.4& 72.9& 61.5& 20.6& 85.8& 92.7& 30.2& 27.3& 27.7& 23.6& 83.2& 64.5& 69.3& 50.1& 45.8& 48.2& 55.2& 41.8& 3.9\\
& LaserMix~\hfill\cite{kong2023lasermix}& 50.6& -& -& -& -& -& -& -& -& -& -& -& -& -& -& -& -& -& -& -\\
& Ours($ w/o~\mathcal{L}_{m\mathcal{I}-\mathrm{FL}}$)~\hfill& 52.7& 79.5& 91.7& 86.4& 71.9& 64.2& 58.0&  87.6& 67.1& 58.1&  27.8& 48.6&  29.2& 22.3& 51.9& 41.3& 35.9&  31.8& 42.6& 5.4\\
& Ours& 54.6& 94.7& 31.1& 39.7& 34.4& 24.5& 51.1& 48.9& 15.3& 90.8& 63.6& 74.1& 47.9& 90.7& 61.5& 82.7& 62.1& 67.5& 51.4& 5.3 \\
\midrule
\midrule
\multirow{4}{*}{1pt} 
&PSD*~\hfill\cite{zhang2021perturbed}& 41.9& 69.9& 89.6& 76.2& 69.1& 61.6& 42.3& 82.7& 53.3& 36.8& 19.1& 23.8& 15.8& 6.1& 46.8& 26.2& 21.7& 22.2& 32.5& 0.4 \\
& HybridCR*~\hfill\cite{li2022hybridcr}& 43.6& 75.3& 21.9& 39.7& 23.6& 26.7& 41.8& 42.1& 20.1& 80.8& 53.1& 45.9& 19.7& 68.2& 61.2& 79.2& 51.1& 42.3& 33.2& 2.5\\
& Ours($ w/o~\mathcal{L}_{m\mathcal{I}-\mathrm{FL}}$)~\hfill& 45.8& 70.5& 89.1& 78.4& 65.9& 60.5& 51.2& 81.0& 59.0& 53.9& 20.2& 38.4& 19.3& 13.0&  48.9& 33.1& 23.1&  29.5& 31.8& 3.4\\
& Ours& 46.5& 73.7& 20.3& 29.2& 22.1& 21.2& 45.9& 50.4& 19.3& 77.3& 53.9& 60.5& 21.9& 77.3& 56.9& 82.7& 62.1& 67.3& 38.1& 3.4\\
\bottomrule
\end{tabular}}
\end{center}
\caption{Per-class quantitative results on SemanticKITTI~\cite{behley2019semantickitti}.In the per-class columns, righter classes tend to be more tail in table.``*'' denotes results trained by official codes.}\label{semantickitti-tab}
\end{table*}
\begin{table*}[htp]
\begin{center}
\resizebox{\textwidth}{17mm}{
\begin{tabular}{c|l|c|c|cccccccc}
\toprule
Set. & Methods & mIoU(\%) & OA& buildings& high-veg.& man-made.& natural.&  low-veg.&  hard-scape & scanning-art.& cars  \\
\midrule
\multirow{7}{*}{Fully}  
&ShellNet~\hfill\cite{zhang2019shellnet}& 69.3& 93.2& 94.2& 83.9& 96.3& 90.4& 41.0& 34.7& 43.9& 70.2\\
&KPConv~\hfill\cite{thomas2019kpconv}& 74.6& 92.9& 94.9& 84.2& 90.9& 82.2& 47.9& 40.0& 77.3& 79.7 \\
&RandLA-Net~\hfill\cite{hu2019randla}& 77.4& 94.8& 95.7& 86.6& 95.6& 91.4& 51.5& 51.5& 69.8& 76.8  \\
&PointGCR~\hfill\cite{ma2020global}& 69.5& 92.1& 93.2& 64.4& 93.8& 80.0& 66.4& 39.2& 34.3& 85.3\\
&RFCR~\hfill\cite{gong2021omni} & 77.8& 95.0& 95.0& 85.7& 94.2& 89.1& 54.4& 43.8& 76.2& 83.7 \\
& HybridCR~\hfill\cite{li2022hybridcr}& 77.4& 95.1& 97.3& 84.1& 87.7& 58.2& 95.2& 48.2& 67.5& 81.0\\
& Ours($ w/o~\mathcal{L}_{m\mathcal{I}-\mathrm{FL}}$)~\hfill& 76.4& 94.8& 97.8& 91.1& 90.3& 54.5& 92.3& 47.2& 66.1& 71.9\\
& Ours& 77.9& 95.1& 98.3& 94.3& 89.3& 61.4& 91.7& 56.7& 59.1& 72.4\\
\midrule
\midrule
\multirow{4}{*}{1\%} 
& PSD~\hfill\cite{zhang2021perturbed}&  75.8& 94.3& 95.1& 86.7& 97.1& 91.0& 48.1& 46.5& 63.2& 79.0\\
& HybridCR~\hfill\cite{li2022hybridcr}& 76.8& 94.9& 97.8& 94.0& 86.6& 52.9& 95.3& 47.1& 64.9& 75.5\\
& Ours($ w/o~\mathcal{L}_{m\mathcal{I}-\mathrm{FL}}$)~\hfill& 76.1& 94.6& 94.2& 84.2& 97.8&  87.6& 48.3& 46.5& 70.4& 79.8\\
& Ours& 76.9& 94.9& 95.2& 85.4& 97.7& 94.6& 53.3& 45.4& 61.2& 82.4\\
\midrule
\midrule
\multirow{4}{*}{1pt} 
&PSD*~\hfill\cite{zhang2021perturbed}&  61.5& 90.8& 80.2& 80.6& 94.0& 76.2& 29.2& 25.5& 39.0& 67.3\\
& HybridCR*~\hfill\cite{li2022hybridcr}&63.5& 91.4& 85.8& 81.1& 79.5& 54.5& 72.2& 37.2& 46.1& 51.6\\
& Ours($ w/o~\mathcal{L}_{m\mathcal{I}-\mathrm{FL}}$)~\hfill& 65.8& 92.1& 84.5& 81.4& 97.2& 76.4& 31.9& 26.2& 58.7& 70.1\\
& Ours& 66.2& 92.8& 88.4& 74.3& 89.3& 66.5& 61.7& 45.7& 51.2& 52.5\\
\bottomrule
\end{tabular}}
\end{center}
\caption{Per-class quantitative results on Semantic3D (reduced-8). OA denotes the overall accuracy of all classes, which is widely-used for evaluating the performance on Semantic3D benchmark~\cite{hackel2017semantic3d}.In the per-class columns, righter classes tend to be more tail in table.``*'' denotes results trained by official codes.}\label{semantic3d-tab}
\end{table*}

\textbf{Implementation details.} 
\uppercase\expandafter{\romannumeral1}-step is trained for 30 epochs to optimize the backbone and \uppercase\expandafter{\romannumeral2}-step is updated for 100 epochs to update the classifier. 
For the training of \uppercase\expandafter{\romannumeral1}-step and \uppercase\expandafter{\romannumeral2}-step, we use Adam Optimizer with an initial learning rate of $0.001$ and momentum of $0.9$ to train $100$ epochs to get the pre-trained model for all datasets on an NVIDIA Titan RTX GPU. The number of neighbor points $K$ is $16$, the batch size is $8$ and the initial learning rate is $0.01$ with a decay rate of $0.98$. 
$\delta_{len}$ and $\delta_{d}$ in Eq.~\ref{eq_pseudo} is set to 0.1 and 0.5, respectively.  
$\beta$ in Eq.~\ref{eq_resample} is $0.5$,
and $s$ in Eq.~\ref{gamma} is $10$.
The iteration times between \uppercase\expandafter{\romannumeral1}-step and \uppercase\expandafter{\romannumeral2}-step is $10$ and the new round of pseudo labels are generated at each iteration. 
Besides, we adopt a point-based backbone (i.e., RandLA-Net~\cite{hu2019randla}) to conduct the experiments. 

\textbf{Evaluation Protocols.}
We evaluate the final performance on all points of the original test set. For the quantitative comparison, we use the mean Intersection-over-Union ($m$IoU) as the standard metric. We experimentally study two types of weak labels: $1$pt and $1\%$ settings. Moreover, we also make comparisons with other state-of-the-art methods in a fully-supervised manner. 
In Tab.~\ref{s3dis_tab}, Tab.\ref{scannet-tab}, Tab.\ref{semantickitti-tab} and Tab.\ref{semantic3d-tab}, decoupling optimization are applied into both $Baseline$ and Ours($\mathcal{L}_{m\mathcal{I}-\mathrm{FL}}$).
$Baseline$ means $\delta_{c}^{\mathrm{cer}}$ is only used to obtain $\hat{Y}$ without $\mathcal{L}_{m\mathcal{I}-\mathrm{FL}}$, \emph{i.e.}, only Eq.~\ref{step1_seg} are used for backbone parameter updating. Ours($\mathcal{L}_{m\mathcal{I}-\mathrm{FL}}$) means both $\delta_{c}^{\mathrm{cer}}$ and $\delta_{c}^{\mathrm{uncer}}$ are used, as well as $\mathcal{L}_{m\mathcal{I}-\mathrm{FL}}$. 

\subsection{Comparison with SOTA Methods}
\textbf{Results on S3DIS.}
First, we compare our method with SOTA methods on S3DIS Area-5 in Tab.~\ref{s3dis_tab}. 
We can observe that our method achieves the highest $m$IoU both in the settings of $1$pt and $1\%$, compared to Zhang~et al..~\cite{zhang2021weakly}, PSD~\cite{zhang2021perturbed}, $\Pi$ Model~\cite{samuli2017temporal}, MT~\cite{tarvainen2017mean}, Xu~et al..~\cite{xu2020weakly}, 
HybridCR~\cite{li2022hybridcr} and GaIA~\cite{lee2023gaia}.
For the $1$pt setting, our method outperforms PSD, HybridCR and GaIA by $6.8\%$, $3.5\%$ and $1.3\%$, respectively. In particular, 
our method achieves $23.1\%$, $16.8\%$ and $8.2\%$ performance gains over PSD in the tail classes of ``board'', ``table'' and ``clutter'', respectively.
Furthermore, for $1\%$ setting, our method achieves $6.4\%$ $m$IoU gains over Zhang~et al.~\cite{zhang2021weakly} and even surpasses Xu~et al.~\cite{xu2020weakly} by $20.2\%$.
Note that S3DIS has rare ``beam'' points, resulting in the test scores of this class being nearly $0.2$ on Area-5.
To explain, our method optimizes the parameters of the network to alleviate the imbalanced issue by effectively decoupling the learning for the classifier and feature with two-stage pseudo-label generation and multi-class imbalanced focal loss. 

\begin{figure*}[t] 
\centering
\includegraphics[width=0.99\textwidth]{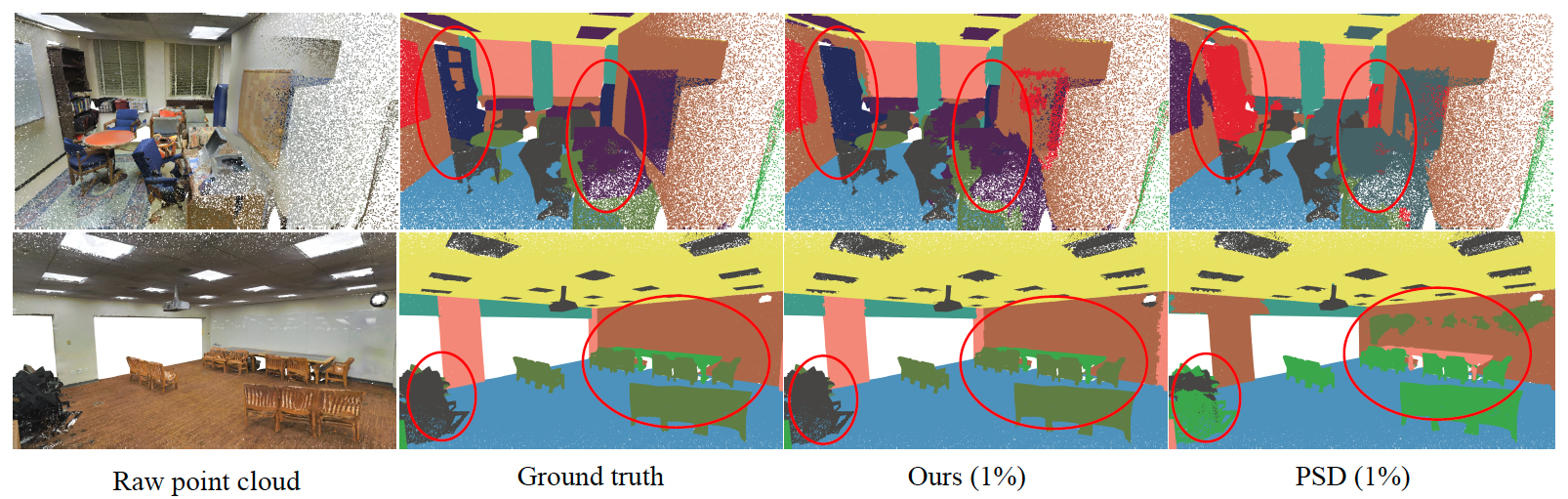}
\caption{Visualization results on the validation set of S3DIS. 
Raw point cloud, semantic labels, ours and results of PSD are presented separately from left to right.
}\label{v_s3dis}
\end{figure*}

As shown in Fig.~\ref{v_s3dis}, we conduct the qualitative comparison on S3DIS. Compared to PSD, our method can generate better segmentation results, especially on ``bookcase'', ``chair'' and ``sofa''. Moreover, our segmentation results are almost consistent with ground-truth segmentation.
To explain, our framework can effectively improve the accuracy of tail classes and promote segmentation performance.

\textbf{Results on ScanNet-V2.}
To further evaluate the effectiveness of our method, we also make a comparison with SOTA methods on ScanNet-V2, as shown in Tab.~\ref{scannet-tab}, our method achieves $67.0\%$ mIoU at the $1\%$ setting, which outperforms GaIA by $1.8\%$, and even surpasses fully-supervised PCNN~\cite{atzmon2018point} by $18.6\%$. For the 1pt setting, our method also achieves $3.6\%$ mIoU gains over GaIA. 
For the evaluation of per-class performance, our method achieves $18.2\%$ and $19.3\%$ mIoU improvements at the setting of $1\%$  on the tail classes ``toilet'' and  ``shower-curtain'' against HybirdCR, respectively.
At the setting on $1pt$, our method achieves $22.2\%$ and $11.1\%$ mIoU gains over HybirdCR on the waist classes ``toilet'' and ``sink'', respectively.
For head classes of ``wall'' and ``chair'' , our method achieves $5.9\%$ and $7.7\%$ mIoU gains comparable performance to HybirdCR.
Our method achieves $1.8\%$ $m$IoU gains over GaIA. at the same annotation setting of $1\%$.

\begin{figure*}[t] 
\centering
\includegraphics[width=0.99\textwidth]{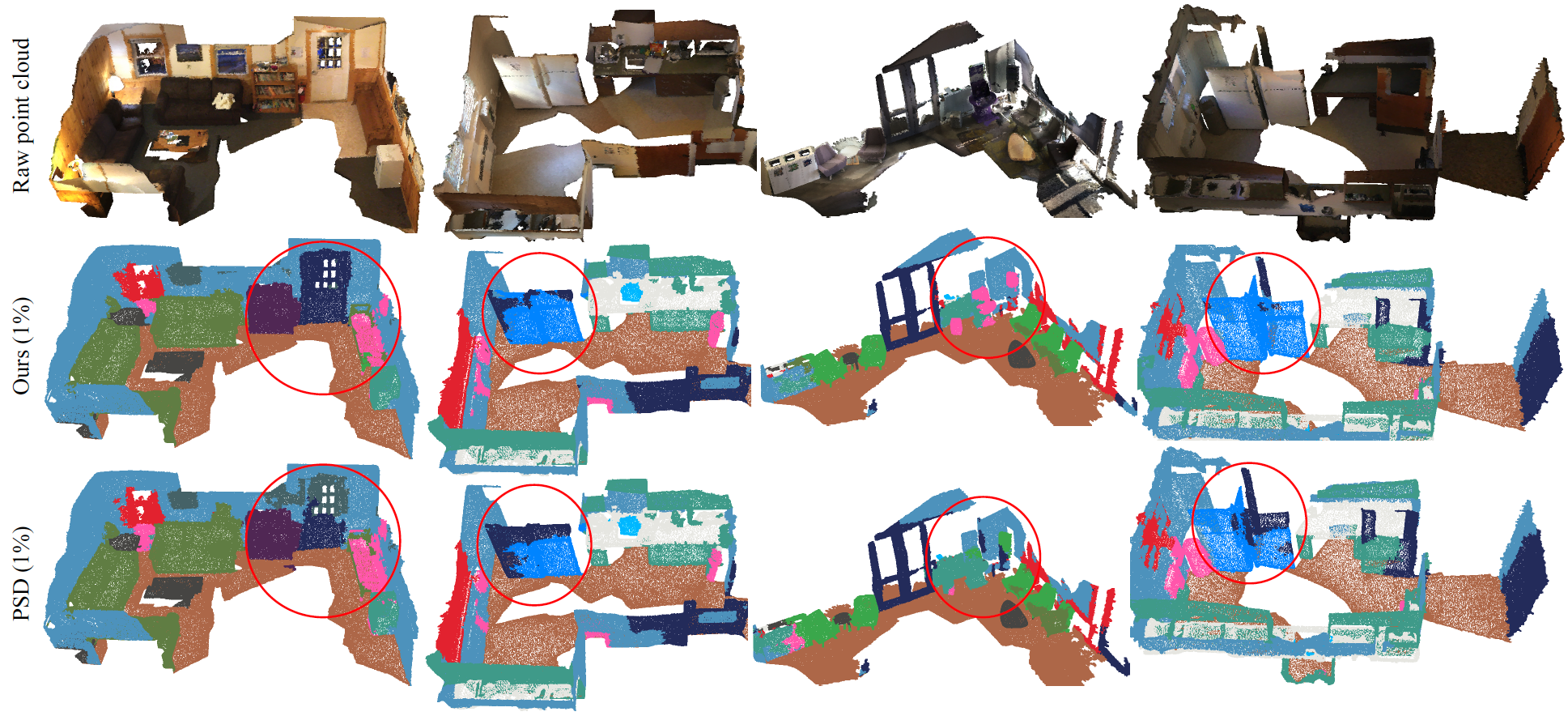} 
\caption{Visualization results on ScanNet-V2.}\label{scannet_fig}
\end{figure*}

As shown in Fig.~\ref{scannet_fig}, we conduct the qualitative comparison on S3DIS. Since there is no public ground truth, we show the raw point clouds at the top row and our segmentation results at the bottom row.
It can be observed that our method can achieve the better segmentation results at the $1\%$ setting, compared to PSD. In particular, the segmentation for the corners and boundaries in the class `` wall'' and ``door'' are more accurate, compared to PSD. 
%
%
\begin{figure*}[t] 
\centering
\includegraphics[width=0.99\textwidth]{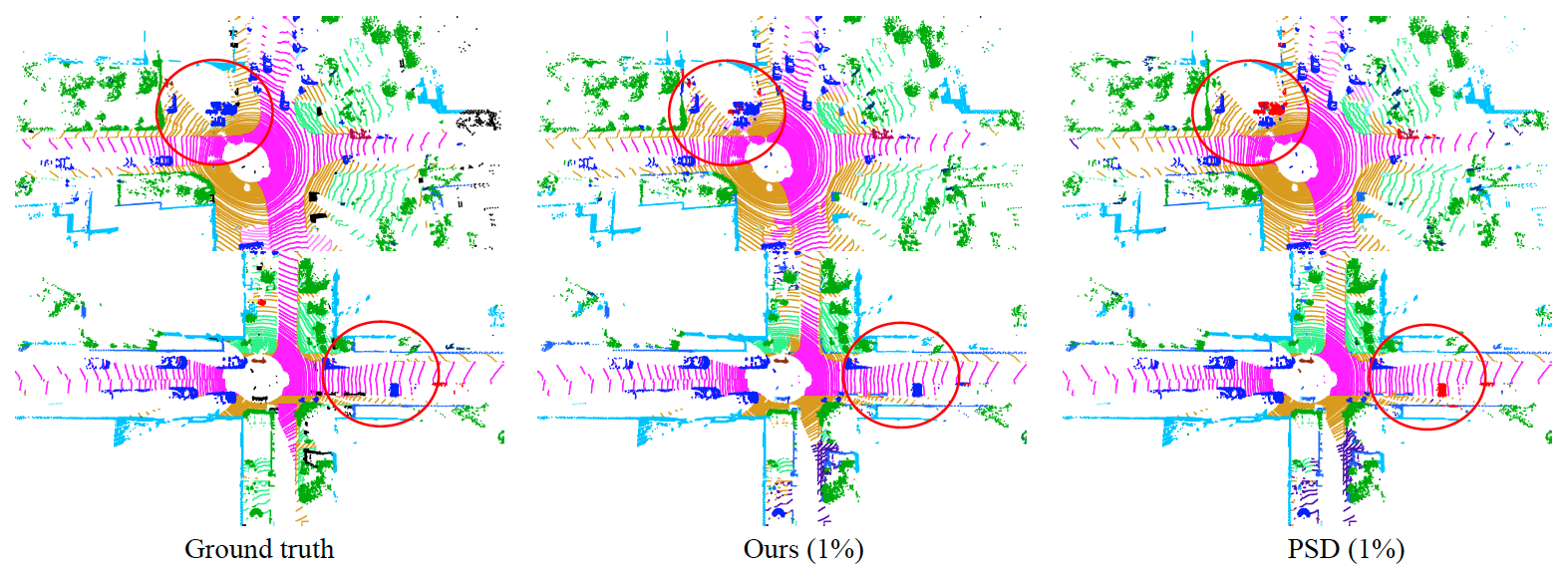}
\caption{Visualization results on the validation of SemanticKITTI. Semantic labels, ours and results of PSD are presented separately from left to right.
}
\label{v_semantickitti}
\end{figure*}

\textbf{Results on SemanticKITTI.}
Then, we conduct the per-class quantitative evaluations on SemanticKITTI, as shown in Tab.~\ref{semantickitti-tab},  
our method achieves the best performance of $57.5\%$ at the fully-supervised setting, compared to PointNet~\cite{qi2017pointnet},SqueezeSegV2~\cite{wu2019squeezesegv2}, DarkNet53Seg~\cite{behley2019semantickitti}, RangeNet53++~\cite{milioto2019rangenet++} and RandLA-Net~\cite{hu2019randla}.
Compared to HybirdCR, our method also achieves the better performance of $54.6\%$ and $46.5\%$ mIoU at the $1\%$ and $1pt$ settings, respectively. 
Our method only labels $1\%$ points even surpassing the fully-supervised RandLA-Net by $0.7\%$ mIoU as well as surpassing latest LaserMix~\cite{kong2023lasermix} by $4.0\%$
For the per-class performance at the $1\%$ setting, we surpass HybirdCR by $9.6\%$ and $1.4\%$ mIoU on the tail classes ``bicyclist'' and ``motorcyclist'', respectively.
At the 1pt setting, our method outperforms HybirdCR by $21.4\%$ and $14.6\%$ on the waist classes ``trunk'' and ``pole''.
For the head classes of ``road'', our method still keeps comparable performance to PSD. 
Besides, we achieve the best performance in the ``parking'' and ``bycicle'' classes.
Therefore, the results demonstrate that our method has reliable performance especially on the tail classes on the outdoor dataset. 
For SemanticKITTI, our method surpasses HybirdCR with $2.9\%$ at $1$pt setting on the test dataset.
Therefore,, the results show that our method can generate to the sparse outdoor dataset and improve the performance of tail classed by a large margin.

As shown in Fig.~\ref{v_semantickitti}, we present the qualitative results on SemanticKITTI. We find that our method achieves consistent segmentation results to ground-truth labels, especially on the tail class of ``car''. 

\textbf{Results on Semantic3D}
We further conduct the per-class quantitative evaluations on Semantic3D (reduced-8), as shown in Tab.~\ref{semantic3d-tab}. Overall Accuracy (OA) of all classes is used as the standard metric on the Semantic3D\cite{hackel2017semantic3d}, as well as mIoU.
We first compare our method with fully-supervised ones, such as ShellNet~\cite{zhang2019shellnet}, KPConv~\cite{thomas2019kpconv}, RandLA-Net~\cite{hu2019randla}, PointGCR~\cite{ma2020global}, RFCR~\cite{gong2021omni} and HybridC~\cite{li2022hybridcr}. 
We found that our method achieves the decrease of only $0.9\%$ and $0.5\%$ in OA and $0.1\%$ and $0.2$ in mIoU using $1\%$ labeled points, compared to RFCR and HybirdCR trained on the fully labeled data.
At the $1pt$ setting, our method outperforms HybirdCR on all the classes. For example, compared to HybirdCR, we achieves the improvement of $5.1\%$ mIoU on the tail class of ``scanning-art'', $12.0\%$ mIoU on the waist class of ``nature'', as well as $2.6\%$ mIoU on the head class of ``buildings''.
At the $1\%$ evaluation, our method surpasses PSD and HybirdCR by $3.4\%$ and $6.9\%$ on the tail classes of ``scanning-art'' and ``cars'', respectively.
For head class of ``high-veg.'', our method still keeps comparable performance to PSD. To explain, our method employs decoupling optimization between feature presentation and classifier, two-round pseudo label generation and multi-class imbalanced focus loss, which can effectively learn feature presentation of points from head-to-tail classes.

\begin{figure*}[t]
\centering
\includegraphics[width=0.99\textwidth]{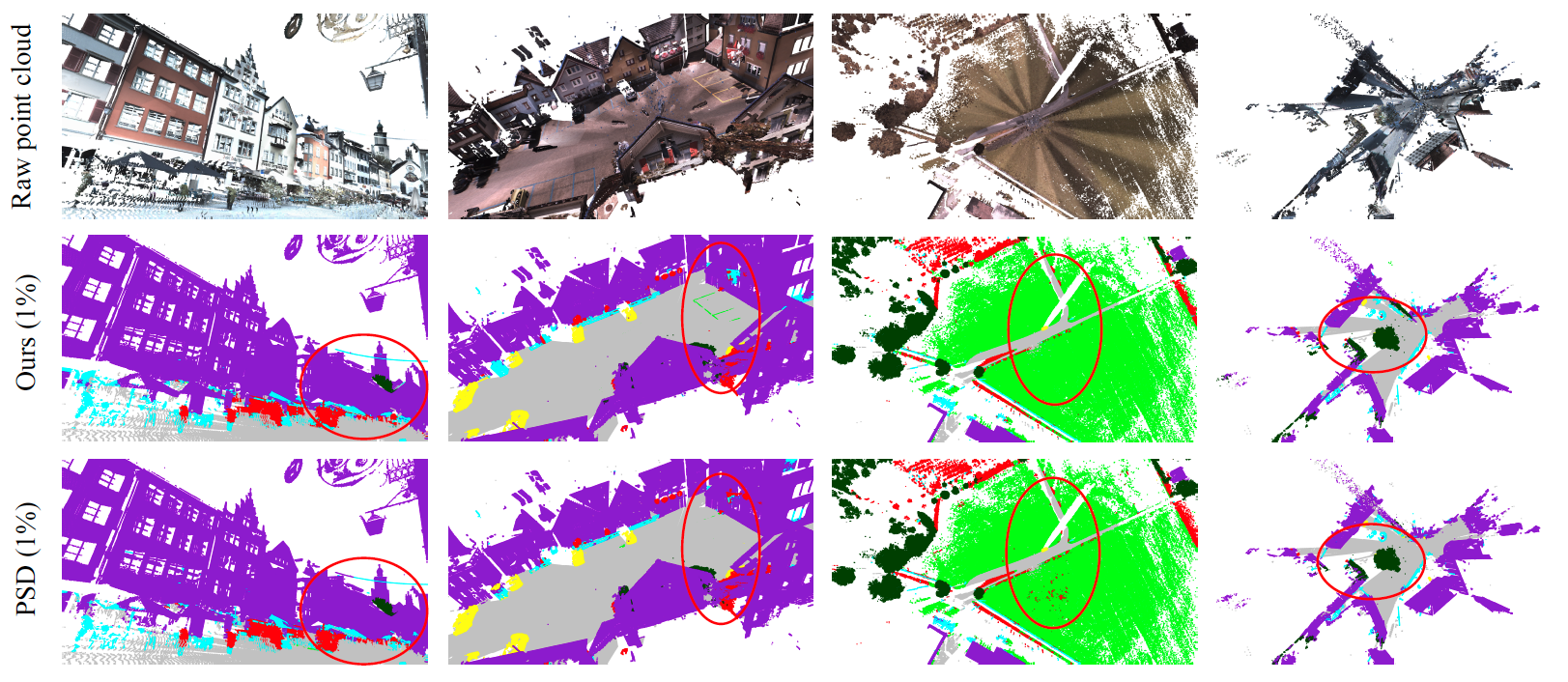} 
\caption{Visualization results on Semantic3D.}\label{semantic3d_fig}
\end{figure*}

Fig.~\ref{semantic3d_fig} shows the visualization results on the test set of Semantic3D. It can be seen that our method can achieve good qualitative segmentation results at the $1\%$ setting. Specifically, our method achieves more accurate predictions for the classes of ``low-veg.'', ``buildings'' and ``man-made'', compared to PSD.

\begin{figure*}[t]
\centering     
\includegraphics[width =0.99\textwidth]{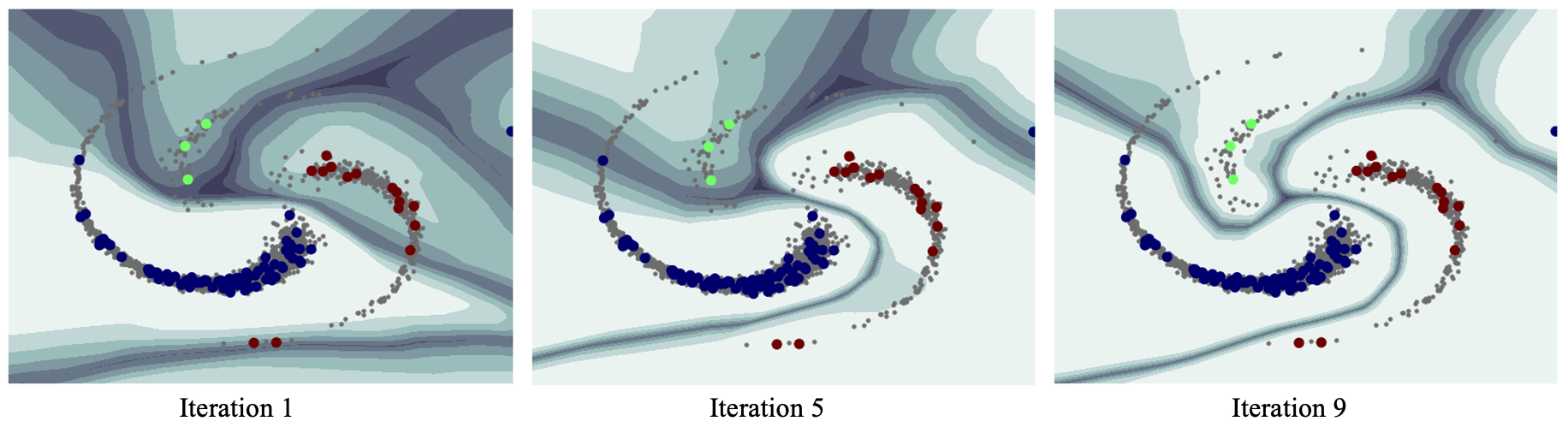}
\caption{Illustration of decision boundaries.  Blue, red and green points refer to ``wall''~(head), ``table''~(waist) and ``sofa''~(tail), respectively. Unlabeled points are denoted by grey colour.}
\label{decision}
\end{figure*}

\textbf{Decision boundaries of Classifier.}
In Fig.~\ref{decision}, we visualize the classifier decision boundaries with the $1^{\mathrm{th}}$ iteration, the $5^{\mathrm{th}}$ iteration and the $9^{\mathrm{th}}$ iteration on S3DIS at $1\%$ setting for three typical head, waist and tail classes with total $\sim$10K points. We could find that with more iterations, the boundaries are more clearly in the feature space of labeled and unlabeled data.
It indicates that during the training, the decision boundaries shift to separate head-to-tail classes without hurting feature generalization.
 \begin{figure*}[t] 
 \centering
 \includegraphics[width=0.98\textwidth]{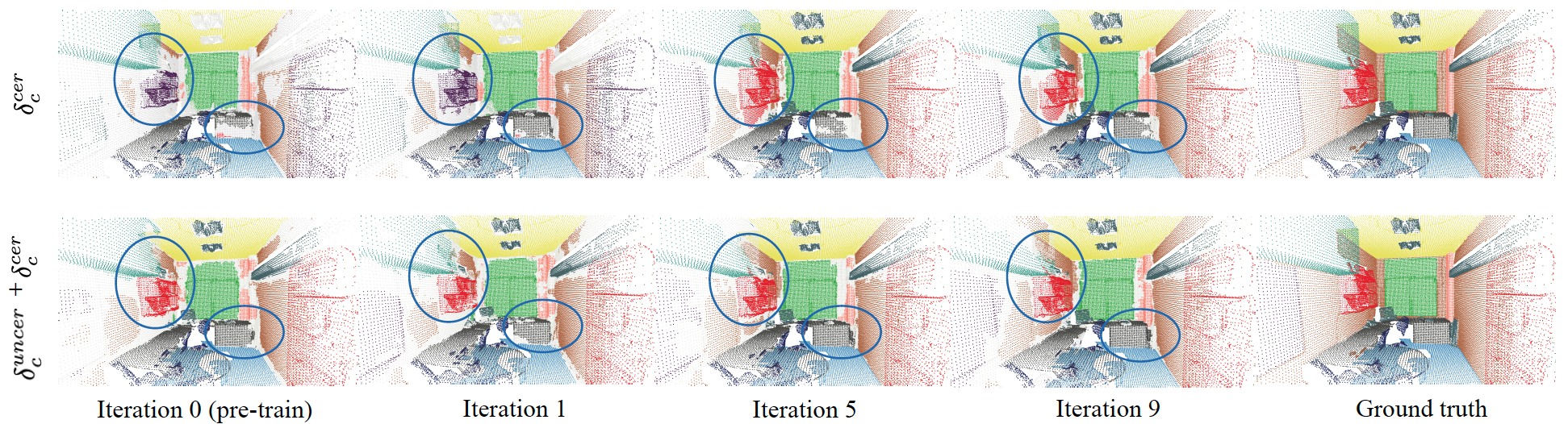}
 \caption{Visualization of pseudo labels generation on the setting about $with$ or $w/o$ $\delta_{c}^{\mathrm{uncer}}$ on four selected iterations.}
 \label{iter_label}
 \end{figure*}
\begin{figure*}[h]
\centering     
\includegraphics[width =0.98\textwidth]{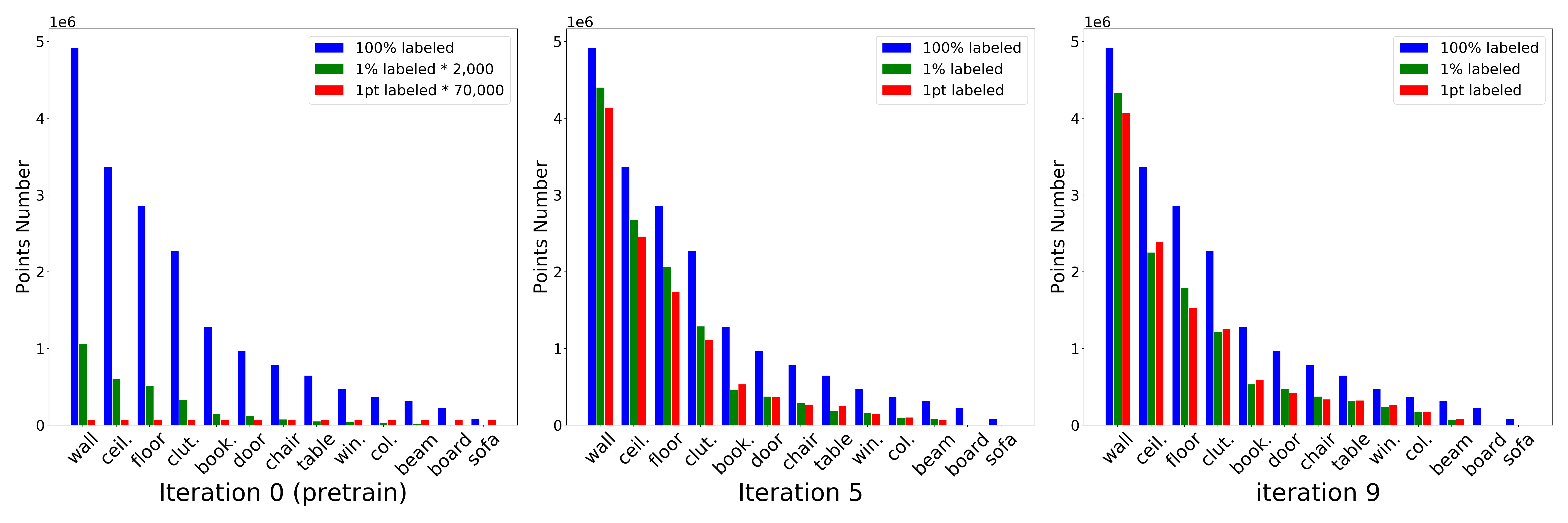}
\vspace{-10pt}
\caption{Per class pseudo labels generation on S3DIS Area-5.}
\label{count}
\end{figure*} 
\subsection{Ablation Study}
We conduct the ablation study on S3DIS Area-5 to evaluate the effect of two-round pseudo-label generation, multi-class imbalanced focus loss, decoupling optimization, and fine-tuning of the classifier.

\textbf{Effect of $\delta_{c}^{\mathrm{uncer}}$.}
\label{ab_resample}
As shown in Tab.~\ref{ab_baseline}, it can be seen from the results (\emph{i.e.}, \#3, \#4) that $\delta_{c}^{\mathrm{uncer}}$ is able to enlarge the pseudo label set with a high probability adding tail classes, which can improve the $m$IoU by $1.5\%$ and $2.0\%$ at $1$pt and $1\%$ setting, respectively.  We visualize the procedure of pseudo labels generation on $1\%$ setting, as shown in Fig.~\ref{iter_label}. We can observe that our strategy provides the correct prediction of class ``board'' in the iteration 0 and 1, and generates more pseudo labels of tail class ''table'' in iterations 5 and 9. 
 Besides, we also visualize the per class count of pseudo labels for three iterations in Fig.~\ref{count}.
 We find that more pseudo labels in the $9^{\mathrm{th}}$ iteration is generated for tail classes compared to that of the $5^{\mathrm{th}}$ iteration in both $1\%$ and $1$pt settings.
%
%

\begin{table*}[t]
\centering
\resizebox{\linewidth}{!}{\begin{tabular}{c|ccc|cc|ccc|cc}
\toprule
 & \multicolumn{3}{c|}{Decouple}                 & \multicolumn{2}{c|}{Pseudo labels}     & \multicolumn{3}{c|}{Optimization(feature learner)}                              & \multicolumn{2}{c}{$m$IoU}         \\ \midrule
 & \multicolumn{1}{c|}{GT} & \multicolumn{1}{c|}{Pseu.} &  \multicolumn{1}{c|}{$ w/o$} &  \multicolumn{1}{c|}{$\delta_{c}^\mathrm{cer}$} &  $\delta_{c}^\mathrm{cer} \& \delta_{c}^\mathrm{uncer}$  & \multicolumn{1}{c|}{$\mathcal{L}_{\mathrm{seg}-\uppercase\expandafter{\romannumeral1}}$} & \multicolumn{1}{c|}{$\mathcal{L}_{\mathrm{FL}}$+ $\mathcal{L}_{\mathrm{seg}-\uppercase\expandafter{\romannumeral1}}$} & $\mathcal{L}_{m\mathcal{I}-\mathrm{FL}}$+$\mathcal{L}_{\mathrm{seg}-\uppercase\expandafter{\romannumeral1}}$ & \multicolumn{1}{c|}{$1$pt} & $1\%$ \\ \midrule
\#1 & \multicolumn{1}{c}{\checkmark}   & \multicolumn{1}{c|}{\checkmark}  &  \checkmark& \multicolumn{1}{c}{\ding{53}}  &  \checkmark & \multicolumn{1}{c}{\checkmark}  & \multicolumn{1}{c}{\ding{53}}  &  \ding{53} & \multicolumn{1}{c|}{48.3}   &  59.8   \\ \midrule
\#2 & \multicolumn{1}{c}{\checkmark}  & \multicolumn{1}{c|}{\ding{53}}  &  \ding{53}&  \multicolumn{1}{c}{\checkmark} & \ding{53}  & \multicolumn{1}{c}{\checkmark}  & \multicolumn{1}{c}{\ding{53}}  &  \ding{53} & \multicolumn{1}{c|}{50.3}   & 61.8     \\ \midrule
\#3 & \multicolumn{1}{c}{\checkmark}  & \multicolumn{1}{c|}{\ding{53}}  &  \ding{53}& \multicolumn{1}{c}{\ding{53}} & \checkmark  & \multicolumn{1}{c}{\ding{53}}  & \multicolumn{1}{c}{\ding{53}}  &  \checkmark & \multicolumn{1}{c|}{55.0}  &  64.2     \\ \midrule
\#4 & \multicolumn{1}{c}{\checkmark}  & \multicolumn{1}{c|}{\ding{53}}  & \ding{53}&  \multicolumn{1}{c}{\checkmark} & \ding{53}  & \multicolumn{1}{c}{\ding{53}}  & \multicolumn{1}{c}{\ding{53}}  &  \checkmark & \multicolumn{1}{c|}{53.5}  &  62.2     \\ 
\midrule
\#5 & \multicolumn{1}{c}{\checkmark}  & \multicolumn{1}{c|}{\ding{53}}  & \ding{53}&  \multicolumn{1}{c}{\ding{53}} & \checkmark  & \multicolumn{1}{c}{\checkmark}  & \multicolumn{1}{c}{\ding{53}}  &  \ding{53} & \multicolumn{1}{c|}{51.9}  & 61.3      \\ 
\#6 & \multicolumn{1}{c}{\checkmark}  & \multicolumn{1}{c|}{\ding{53}}  & \ding{53}& \multicolumn{1}{c}{\ding{53}} & \checkmark  & \multicolumn{1}{c}{\ding{53}}  & \multicolumn{1}{c}{\checkmark}  &  \ding{53} & \multicolumn{1}{c|}{52.3}  & 62.7   \\ 
\midrule
\#7 & \multicolumn{1}{c}{\checkmark}  & \multicolumn{1}{c|}{\checkmark}  &  \checkmark& \multicolumn{1}{c}{\ding{53}}  &  \checkmark & \multicolumn{1}{c}{\ding{53}}  & \multicolumn{1}{c}{\ding{53}}  &  \checkmark & \multicolumn{1}{c|}{51.7}   &   61.5 \\ 
\#8 & \multicolumn{1}{c}{\checkmark}  & \multicolumn{1}{c|}{\checkmark}  &  \ding{53}& \multicolumn{1}{c}{\ding{53}} & \checkmark  & \multicolumn{1}{c}{\ding{53}}  & \multicolumn{1}{c}{\ding{53}}  &  \checkmark & \multicolumn{1}{c|}{55.2}  &  64.0   \\ 
\bottomrule
\end{tabular}}
\caption{Ablations studies on the effect of decoupling optimization, two-round pseudo label generation and multi-class imbalanced focal loss. \#$2$ is Baseline and \#$3$ is Ours~($\mathcal{L}_{m\mathcal{I}-\mathrm{FL}}$).}
\label{ab_baseline}
\end{table*}
\begin{table}
\centering
\begin{adjustbox}{max width=0.7\linewidth}
\begin{tabular}{c|ccc|cc}
\toprule
  & \multicolumn{3}{c|}{Classifier}                           & \multicolumn{2}{c}{$m$IoU}      \\ \midrule
  & \multicolumn{1}{c|}{GT} & \multicolumn{1}{c|}{Pseo.} & GT+Pseo. & \multicolumn{1}{c|}{$1$pt} & $1\%$ \\ \midrule
\#1 & \multicolumn{1}{c|}{\checkmark}   & \multicolumn{1}{c|}{}   &       & \multicolumn{1}{c|}{55.0}    &   64.2 \\ \midrule
\#2 & \multicolumn{1}{c|}{}   & \multicolumn{1}{c|}{\checkmark}   &       & \multicolumn{1}{c|}{54.1}    &   63.8 \\ \midrule
\#3 & \multicolumn{1}{c|}{}   & \multicolumn{1}{c|}{}   &       \checkmark& \multicolumn{1}{c|}{55.2}    &   64.0 \\ 
\bottomrule
\end{tabular}
\end{adjustbox}
\caption{Effect of label sets for fine-tuning the classifier at the same decouple training. \#1 is Ours.}
\label{ab_classifier}  
\end{table}

\begin{figure*}[t]
\centering     
 \includegraphics[width =0.99\textwidth]{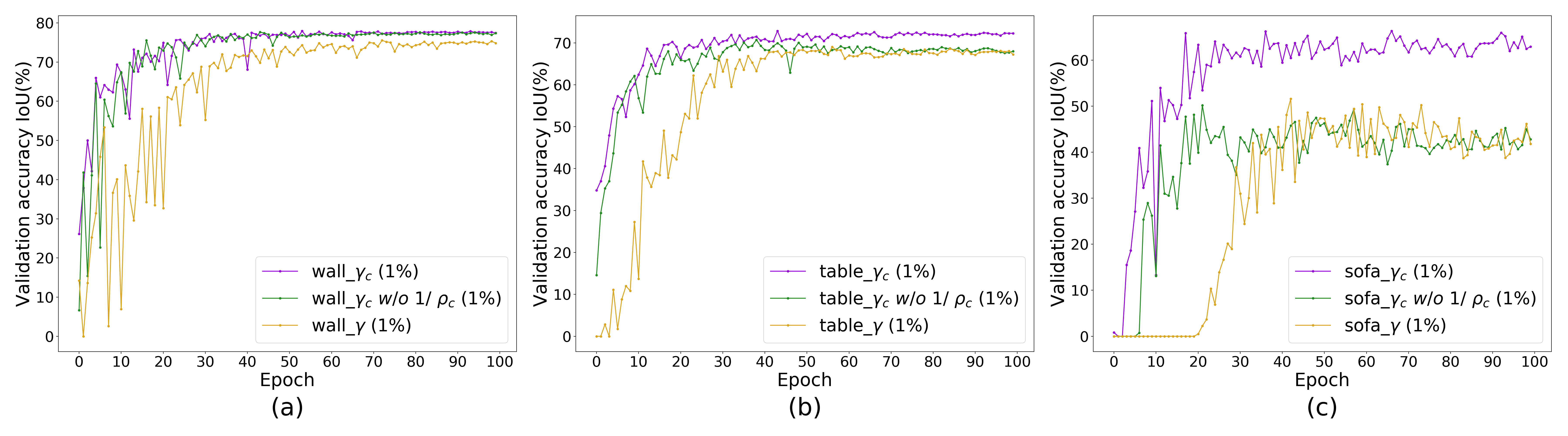}
\caption{Visualization of validation accuracy IoU of $\gamma$ of $\mathcal{L}_{\mathrm{FL}}$,  $\gamma_{c}$ of $\mathcal{L}_{m\mathcal{I}-\mathrm{FL}}$ with and without imbalanced rate  $ 1 /\ \rho_{c}$  at $1\%$ setting on three typical classes: (a) head(``wall''), (b) waist(``table'') and (c) tail(``sofa'').}
\label{epoch_gama}
\end{figure*}

\textbf{Effect of $\mathcal{L}_{m\mathcal{I}-\mathrm{FL}}$.}
\label{ab_mxfl}
As shown in Tab.~\ref{ab_baseline}, we can find the effectiveness of $\mathcal{L}_{m\mathcal{I}-\mathrm{FL}}$ from (\#3, \#5, \#6), which improves the performance of our method by $3.1\%$ and $2.9\%$ at $1$pt and $1\%$ setting, respectively. In Fig.~\ref{epoch_gama}, we further visualize the validation $m$IoU to investigate the effects of the $\gamma$ of $\mathcal{L}_{\mathrm{FL}}$,  $\gamma_{c}$ of $\mathcal{L}_{m\mathcal{I}-\mathrm{FL}}$ with or without imbalanced rate at $1\%$ setting during training.
We select a specific \uppercase\expandafter{\romannumeral1}-step at iteration $2$ and obtain the curve of the head (``wall''), waist (``table'') and tail (``sofa'') classes. We found that $\gamma_{c}$ achieves the best performance, especially on ``soft'', compared to that using $\gamma_{c}$ without imbalanced ratio $\rho_{c}$.
Beside, $s\left(1-g^{c}\right)$ is better than $\gamma$, which indicate that it handles the imbalance problem of each class independently.
\begin{figure*}[t]
\centering     
 \includegraphics[width=0.99\textwidth]{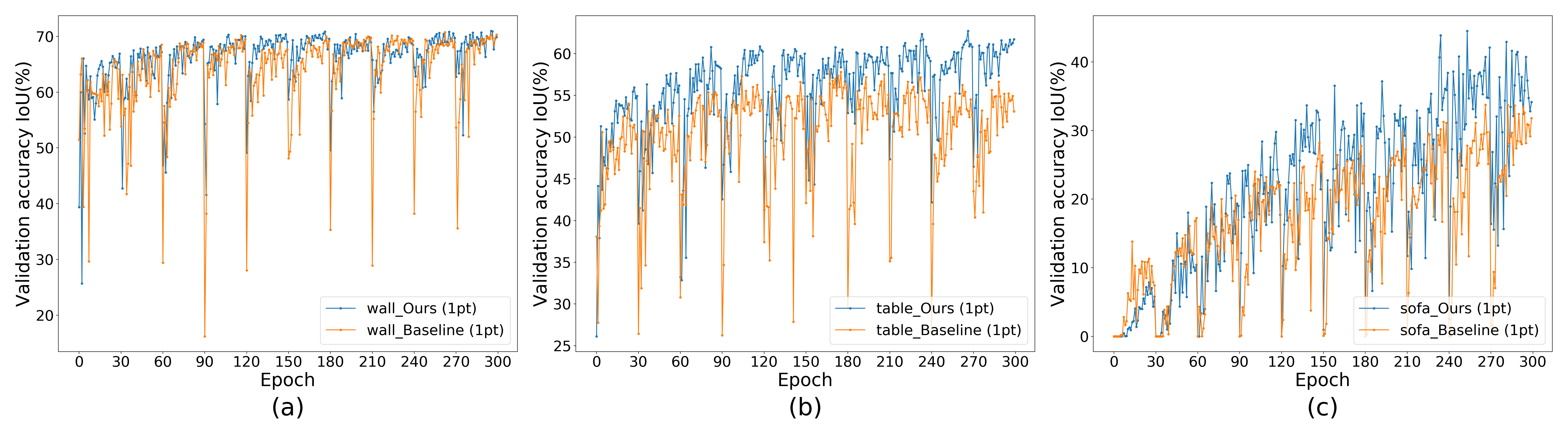}
\caption{Visualization of validation accuracy IoU at $1$pt setting during training on three typical classes: (a) head~(``wall''), (b) waist~(``table'') and (c) tail~(``sofa'').}
\label{class_epoch}
\end{figure*}

\textbf{Effect of used labels for classifier fine-tuning.}
\label{ab_labelsetting}
As shown in Tab.~\ref{ab_classifier}, using points of ground-truth (GT) labels (\emph{i.e.}, \#1) achieves higher $m$IoU compared to that of using only pseudo labels (\emph{i.e.}, \#2).
Mixed GT labels and pseudo labels (\emph{i.e.}, \#3) achieve relatively consistent performance to \#1. This is because the testing set of point clouds follows the long-tail distribution as the same as the training set, while there is a relatively small number of parameters for updating in the classifier. Thus, we adapt the ground truth labels to fine-tune the classifier without reloading the pseudo labels once again in \uppercase\expandafter{\romannumeral2}-step.

\textbf{Effect of decoupling training.}
As shown in Tab.~\ref{ab_baseline}, we investigate the effectiveness of the decouple optimization (\emph{i.e.}, \#8) strategy, compared to joint training (\emph{i.e.}, \#7). We can see that the decouple optimization achieves large margin improvements with $3.5\%$ and $2.5\%$ $m$IoU  at the $1$pt and $1\%$ settings, respectively.
\textbf{The Effect of multi-class imbalanced focus loss in the classifier (\uppercase\expandafter{\romannumeral2}-step).}
\label{cassifier}
As shown in Tab.~\ref{focal-tab}, when the $\mathcal{L}_{m\mathcal{I}-\mathrm{FL}}$ is moved to the learning of classifier in (\uppercase\expandafter{\romannumeral2}-step), we find that the performance increases $2.8\%$ $m$IoU at $1$pt setting. What is more, the performance would be decreased if the $\mathcal{L}_{m\mathcal{I}-\mathrm{FL}}$ is fixed in the (\uppercase\expandafter{\romannumeral1}-step), which shows the effectiveness of our proposed novel loss function and the training settings.
\begin{table}[t]
\begin{center} 
\begin{adjustbox}{max width=1.0\linewidth}
\begin{tabular}{l|l|l}
\hline
\uppercase\expandafter{\romannumeral1}-step & \uppercase\expandafter{\romannumeral2}-step & 1pt \\ \hline
$\mathcal{L}_{m\mathcal{I}-\mathrm{FL}} + \mathcal{L}_{\mathrm{seg}-\uppercase\expandafter{\romannumeral1}}$ & $\mathcal{L}_{\mathrm{seg}-\uppercase\expandafter{\romannumeral2}}$    & 55.0    \\ \hline
$\mathcal{L}_{\mathrm{seg}-\uppercase\expandafter{\romannumeral1}}$    & $\mathcal{L}_{m\mathcal{I}-\mathrm{FL}} + \mathcal{L}_{\mathrm{seg}-\uppercase\expandafter{\romannumeral2}}$ & 52.2   \\ \hline
\end{tabular}
\end{adjustbox}	
\end{center}
\caption{Ablations of loss functions on S3DIS-Area 5.}
\label{focal-tab}
\end{table}
 \begin{table}[htb]
\begin{center}
\begin{adjustbox}{max width=1.\linewidth}
\begin{tabular}{ccc|ccc|cc}
\hline
\multicolumn{3}{c|}{ $\beta$}                                    & \multicolumn{3}{c|}{$\delta_{d}$}                                    & \multicolumn{2}{c}{1pt}               \\ \hline
\multicolumn{1}{c|}{0.2} & \multicolumn{1}{c|}{0.5} & 0.8 & \multicolumn{1}{c|}{0.4} & \multicolumn{1}{c|}{0.5} &           0.6 & \multicolumn{1}{c|}{$\delta_{len}$=0.1} & $\delta_{len}$=0.2  \\ \hline
\checkmark &                          &     &              \checkmark  &                          &               & \multicolumn{1}{c|}{53.4}        &   52.7     \\ \hline
\checkmark &                          &     &                          &              \checkmark  &               & \multicolumn{1}{c|}{54.3}        &    53.1    \\ \hline
\checkmark &                          &     &                          &                          &   \checkmark  & \multicolumn{1}{c|}{53.9}        &   54.9     \\ \hline
&             \checkmark   &     &              \checkmark  &                          &               & \multicolumn{1}{c|}{54.2}        &   53.3     \\ \hline
&             \checkmark   &     &                          &              \checkmark  &               & \multicolumn{1}{c|}{55.0}    &   54.5     \\ \hline
&             \checkmark   &     &                          &                          &   \checkmark  & \multicolumn{1}{c|}{54.7}        & 54.1   \\ \hline
&                          & \checkmark &       \checkmark  &                          &               & \multicolumn{1}{c|}{54.1}        &    53.9   \\ \hline
&                          & \checkmark &                   &              \checkmark  &               & \multicolumn{1}{c|}{54.6}        &     54.2    \\ \hline
&                          & \checkmark &                   &                          &   \checkmark  & \multicolumn{1}{c|}{54.8}        &     54.5    \\ \hline
\end{tabular}
\end{adjustbox}	
\end{center}
\caption{Ablation study of $\beta$, $\delta_{d}$ and $\delta_{len}$ on S3DIS Area-5.}
\label{hyper-tab}
\end{table}
\begin{table}[t]
\begin{center}
\begin{adjustbox}{max width=1.0\linewidth}
\begin{tabular}{l|cccc}
\hline
Method& \makecell[c]{Training \\time(s)}& \makecell[c]{Network \\parameters(M)}& \makecell[c]{Total reference \\time(s)}& \makecell[c]mIoU(\%) \\
\hline
RandLA-Net& 216& 1.05& 258& 62.4\\
PSD~\hfill(1\%)& 302& 1.10& 263& 65.1\\
Ours~\hfill(1\%)& 221(\uppercase\expandafter{\romannumeral1}-step)+103(\uppercase\expandafter{\romannumeral2}-step) & 1.06& 251& 66.6 \\
\hline
\end{tabular}
\end{adjustbox}	
\end{center}
\caption{Model complexity running all points on S3DIS Area-5.}
\label{model-tab}
\end{table}

\textbf{Ablation for hyper-parameters $\beta$, $\delta_{d}$, $\delta_{len}$ and $s$.}
\label{hyper_para}
In Tab.~\ref{hyper-tab}, we conduct ablation studies of these hyper-parameters, where $s$ is set to $10$ by empirically following EQLv2~\cite{tan2021equalization}. We find that the default setting of $\beta(=0.5)$, $\delta_{d}(=0.5)$, $\delta_{len}(=0.1)$ achieves the best $m$IoU.

\textbf{Comparison on the model complexity.}
\label{model_complexity}
 As shown in Tab.~\ref{model-tab}, our method requires relatively similar inference time and parameter number, while we achieve the highest \textit{m}IoU, compared to RandLA-Net and PSD.
\subsection{The Entire Training from Head-to-tail Classes}
As shown in Fig.~\ref{class_epoch}, we conduct the 10-iteration training process for our method to obtain the performance change of three typical classes of S3DIS on the head~(``wall''), waist~(``table'') and tail~(``sofa'') classes.
Compared to baseline, \emph{i.e.}, Ours($ w/o~\mathcal{L}_{m\mathcal{I}-\mathrm{FL}}$), ours is able to improve the performance on the waist and tail classes, and will not reduce the performance on the head classes. This also demonstrates that our method is more effective in handling complex 3D point cloud class-imbalanced problems when labeling a tiny fraction of labeled points. More results on the total 13 classes at $1\%$ and $1$pt settings are presented in the supplementary material.

%% file: latex/conclusion.tex
\section{Conclusion}
\label{conclusion}
In this paper, we propose a new decoupling optimization framework for imbalanced SSL on large-scale 3D point clouds.
It decouples the learning of the backbone and classifier in an alternative optimization manner, which can effectively shift the bias decision boundary to achieve high performance. The parameters of the backbone are updated
by the proposed two-round pseudo-label generation and multi-class imbalanced focus loss, while the classifier is simply fine-tuned using ground-truth data.
Extensive experiments on indoor and outdoor datasets demonstrate the effectiveness of our proposed method, which outperforms the previous SOTA methods at $1\%$ and $1$pt settings.
On S3DIS Area-5, we surpass PSD and Zhang~et al. by $6.8\%$ and $2.4\%$ at $1$pt and $1\%$ settings, respectively.

